\documentclass[lettersize,journal]{IEEEtran}
\usepackage{amsmath,amsfonts}
\usepackage{algorithmic}
\usepackage{algorithm}
\usepackage{array}
\usepackage[caption=false,font=normalsize,labelfont=sf,textfont=sf]{subfig}
\usepackage{textcomp}
\usepackage{stfloats}
\usepackage{url}
\usepackage{verbatim}
\usepackage{graphicx}
\usepackage{cite}
\usepackage{multirow}
\usepackage{makecell}
\begin{document}

\title{A Grid-Based Framework for E-Scooter Demand Representation and Temporal Input Design for Deep Learning: Evidence from Austin, Texas}

\author{Mohammad Sahnoon$^{1}$,~\IEEEmembership{Member,~IEEE,}  Merkebe Getachew Demissie$^{2}$,
and Roberto Souza$^{3}$,~\IEEEmembership{Member,~IEEE} %

\thanks{*This work was not supported by any organization.}%
\thanks{$^{1}$Mohammad Sahnoon is a Ph.D. candidate at the Department of Electrical and Software Engineering, Schulich School of Engineering, University of Calgary, Calgary, AB T2N 1N4, Canada
        {\tt\small mohammad.sahnoon@ucalgary.ca}}%
\thanks{$^{2}$Merkebe Getachew Demissie is with the Department of Civil Engineering, Schulich School of Engineering, University of Calgary, Calgary, AB T2N 1N4, Canada
        {\tt\small merkebe.demissie@ucalgary.ca}}%
\thanks{$^{3}$Roberto Souza is with the Department of Electrical and Software Engineering, Schulich School of Engineering, University of Calgary, Calgary, AB T2N 1N4, Canada
        {\tt\small roberto.souza2@ucalgary.ca}}%
}

\markboth{ArXiv Preprint}%
{Sahnoon \MakeLowercase{\textit{et al.}}: Image-based E-scooter Demand Prediction}

\IEEEpubid{0000--0000/00\$00.00~\copyright~2021 IEEE}

\maketitle
\begin{abstract}
Despite rapid advances in deep learning for shared micromobility demand prediction, the systematic design and statistical validation of temporal input structures remain underexplored. Temporal representations are often defined heuristically, even though historical demand input strongly influences model performance and generalizability. To address this gap, this paper develops a reproducible data-processing pipeline and a statistically grounded methodology for designing temporal input structures for image-to-image demand prediction. Using large-scale shared micromobility data from Austin, Texas, we construct a grid-based spatiotemporal e-scooter dataset by transforming trip records into hourly pick-up and drop-off demand images on a uniform spatial grid. The pipeline includes trip-record filtering and validation, geospatial mapping of census tract identifiers to spatial locations, grid construction, demand aggregation, and generation of a global activity mask restricting evaluation to locations with historically active micromobility demand. This representation enables consistent spatial learning while preserving demand patterns across the study area. Building on this dataset framework, we introduce a combined correlation- and error-based procedure to identify informative historical demand patterns. The optimal temporal depth is determined through a controlled ablation study using a baseline UNET model with paired non-parametric testing and Holm correction. The resulting temporal configurations capture short-term persistence and daily and weekly periodicities. Compared with adjacent-hour and fixed-period baselines, the proposed temporal input design reduces mean squared error by up to 37\% for next-hour prediction and 35\% for next 24-hour prediction. These findings highlight the importance of principled dataset construction and statistically validated temporal context design for spatiotemporal micromobility demand prediction.
\end{abstract}


\section{Introduction}
\IEEEPARstart{S}{hared} micromobility systems have emerged as a transformative component of urban transportation, offering flexible, scalable, and environmentally sustainable mobility options that complement public transit and private vehicles \cite{Machado2018OverviewShared, Shaheen2024BenefitsCarpooling}. These services, including bike-sharing and dockless scooter-sharing, address the persistent first-mile and last-mile problem \cite{Gonzalez2023UtilizationRate} and contribute to reduced congestion, lower emissions, and increased travel equity \cite{Teusch2023SystematicLiterature, Beza2025EquityImplications, Beojone2021InefficiencyRide, Hollingsworth2019AreE}. As the adoption of shared micromobility continues to grow worldwide, effective operational management becomes increasingly dependent on accurate demand prediction \cite{Teusch2023SystematicLiterature}.

Among shared micromobility modes, dockless electric scooters (e-scooters) have experienced particularly rapid growth due to their convenience and accessibility. Unlike traditional station-based systems, free-floating e-scooters can be picked up and dropped off at any location, which introduces unique challenges for spatial demand modelling \cite{Kim2022PredictingDemand}. Accurate spatiotemporal prediction of e-scooter pick-up and drop-off demand is essential for fleet management, rebalancing strategies, infrastructure planning, and user service level assurance \cite{Baumgarte2021YouLl}.

Despite the critical importance of micromobility demand prediction, the existing literature exhibits several methodological limitations. First, while several studies adopt grid- or zone-based spatial representations for demand modelling \cite{Ham2021SpatiotemporalDemand, Kim2022PredictingDemand}, the transformation from raw trip records to gridded demand surfaces is typically presented as a preprocessing step rather than as a formally defined and reproducible methodological framework. In some cases, key design choices such as aggregation strategy and treatment of sparse demand regions are not systematically articulated, despite their direct impact on data sparsity, spatial bias, and model learning stability.

Second, although recent studies employ advanced recurrent or attention-based architectures, the historical input horizon is typically defined as a fixed sliding window or a pre-specified look-back period selected heuristically or tuned empirically, rather than derived through statistically grounded time-lag ranking or formal hypothesis testing \cite{Ham2021SpatiotemporalDemand, Kim2022PredictingDemand, Xu2023RealTimeForecasting}. While such models can learn temporal dependencies within the provided window, the choice of window length itself is rarely justified through systematic statistical validation, potentially overlooking informative periodic or non-adjacent temporal structures.

Third, commonly used evaluation practices frequently rely on aggregate global error metrics computed over the entire spatial domain \cite{Ham2021SpatiotemporalDemand, Kim2022PredictingDemand}, without incorporating spatial masking strategies that isolate regions with observed historical demand.\IEEEpubidadjcol In sparse grid settings characteristic of dockless e-scooter systems, averaging errors over large zero-demand areas can dilute performance interpretation and obscure differences in operationally relevant zones.

To address these gaps, this paper develops a principled framework for spatiotemporal e-scooter demand prediction. The main contributions of this work are threefold. First, we develop a formalized data-processing pipeline that systematically transforms raw micromobility trip records into a structured spatiotemporal dataset consisting of hourly pick-up and drop-off demand images constructed over a fixed urban grid. The pipeline includes filtering and validation of trip records, geospatial mapping of census tract identifiers to spatial locations, spatial discretization through grid construction, temporal aggregation of trips into hourly demand maps, and the generation of a global binary activity mask that restricts learning and evaluation to spatial regions with historically active demand. The resulting dataset spans the full 2019 calendar year of shared e-scooter trip activity in Austin, TX. Second, building upon this structured representation, we propose a combined correlation- and error-based analysis to identify the most informative historical time lags for demand prediction. Third, we determine the appropriate temporal depth through a controlled ablation study using paired non-parametric hypothesis testing with Holm correction, enabling statistically reliable model selection. Experiments using large-scale e-scooter trip data from Austin, TX demonstrate that the resulting temporal input design yields statistically meaningful prediction improvements over commonly used fixed heuristic baselines.

The remainder of the paper is organized as follows. Section \ref{sec:related-work} reviews related work on shared micromobility demand prediction and existing methodological practices. Section \ref{sec:data} presents the dataset description and the spatiotemporal processing pipeline, including the transformation from raw trip records to grid-based demand images. Section \ref{sec:exp-design} describes the experimental setup, including the prediction horizons, temporal dataset splitting strategy, input representation and spatiotemporal channel construction, deep learning model, loss function, evaluation metrics, and statistical testing protocol. Section \ref{sec:temporal_input_design} introduces the proposed temporal lag identification and depth selection methodology. Section \ref{sec:results} provides empirical evaluation and comparative analysis of the proposed temporal input configurations. Section \ref{sec:discussion} discusses the findings and their implications. Finally, Section \ref{sec:conclusion} concludes the study with key insights and directions for future research.

\section{Related Work}
\label{sec:related-work}

\noindent Recent advances in micromobility demand prediction have increasingly focused on deep learning architectures to capture complex spatiotemporal demand patterns in urban environments. These studies leverage graph-based, encoder–decoder, and attention-based architectures to better capture heterogeneous urban structure and spatial dependencies. Song et al. \cite{Song2023SparseTrip} developed a sparse trip demand prediction framework based on a spatiotemporal graph neural network that integrates built-environment, weather, and periodic features, reporting substantial gains over the graph convolutional recurrent network (GCRN) baseline for hourly e-scooter demand in Louisville, Kentucky. Our prior study \cite{Sahnoon2024UNETUNETR} examined UNET \cite{UNet2015Convolutional} and UNET transformer (UNETR) \cite{Hatamizadeh2022UnetrTransformers} encoder-decoder frameworks against the masked fully convolutional network (MFCN) \cite{Phithakkitnukooon2021PredictingSpatiotemporal} for short-term spatiotemporal e-scooter demand prediction using the Calgary e-scooter dataset under the same experimental settings. Within that comparison, UNETR achieved the lowest mean absolute errors for next-hour and next 24-hour demand prediction, while UNET provided superior accuracy at nonzero-demand locations; both architectures outperformed the MFCN baseline across most demand levels. These findings demonstrated the effectiveness of image-based encoder-decoder models for e-scooter demand prediction and motivate the present study, which builds on that modelling foundation by focusing on the construction of a principled spatiotemporal dataset pipeline and the statistically grounded design of historical temporal input.

Beyond architectural advances, spatial representation has become a central component of micromobility demand modelling. Region-based and grid-based approaches are widely adopted to capture fine-grained spatial heterogeneity while preserving structured model input. For example, Yang et al. \cite{Yang2023RelationshipBetween} divided the urban area into $500\,m\,\times\,500\,m$ grid cells and assigned built-environment features such as POI density, transit accessibility, and network accessibility to each cell, using support vector regression to model shared bicycle usage. On the deep learning side, several methods construct tensor-style or image-style spatial input that combine demographic, functional, or supply-side indicators with historical demand to predict micromobility demand \cite{Phithakkitnukooon2021PredictingSpatiotemporal, Sahnoon2024UNETUNETR, Xu2025ICNInteractive}. A recent contribution \cite{Paul2023EstimatingCensored} further modelled censored spatiotemporal demand for dockless scooter systems over a spatial grid by probabilistically accounting for unobserved demand, demonstrating that grid-level estimation can better reflect usage patterns than na\"ive station- or zone-based aggregation. Compared to these approaches, the present study formalizes a complete dataset processing pipeline that systematically transforms raw trip records into structured hourly demand images through polygon-based census-tract mapping followed by a controlled rectangular grid transformation, providing an explicitly defined spatial representation compatible with image-based deep learning architectures.

In addition to architectural and spatial design considerations, statistical rigor in micromobility demand prediction remains limited. Many studies compare models based solely on point estimates of mean absolute error (MAE), mean squared error (MSE), or root mean square error (RMSE), without assessing whether observed performance differences are statistically significant. Dem\v{s}ar \cite{Demsar2006StatisticalComparisons} provided foundational guidance on non-parametric statistical tests such as the Wilcoxon signed-rank test for comparing multiple machine learning models across common datasets, yet such testing practices remain uncommon in applied deep learning studies. Within the micromobility demand prediction literature, in particular, formal paired hypothesis testing across model configurations is rarely reported. The present work addresses this gap by applying paired non-parametric Wilcoxon signed-rank tests with Holm correction across trained model configurations to determine whether the observed performance improvements are statistically meaningful.

\section{Dataset and Spatiotemporal Processing Pipeline}\label{sec:data}
\noindent Shared micromobility services have played a significant role in Austin's urban mobility landscape since their introduction in early April 2018, when dockless e-scooter fleets first became available to the public. Austin, TX is one of the fastest-growing major cities in the United States, with a population of 961,855 according to the 2020 Decennial Census \cite{Bureau20202020Decennial}. The city also has a relatively young population, with a median age of approximately 34.8 years based on the 2024 American Community Survey 1-Year Estimates \cite{Bureau2024AmericanCommunity}. This relatively young and rapidly expanding urban population presents a compelling setting for micromobility adoption and usage analysis. Major operators such as Bird and Lime deployed thousands of vehicles across the city, and during the early months of operation substantial ridership was recorded. Between September 5 and November 30, 2018, a total of 936,110 e-scooter trips were taken, accounting for 182,333 hours of use and 891,121 miles (1,434,120.23 kilometers) traveled \cite{Health2019DocklessElectric}. According to City of Austin Shared Mobility Services portal \cite{Austinn.d.SharedMobility}, Lime operates approximately 3,700 e-scooters and 180 e-bikes in the city, while Bird operates about 3,000 e-scooters. The dataset used in this study originates from the City of Austin's open-data reporting framework, established to monitor system performance, usage patterns, and operational compliance, and constitutes a comprehensive citywide source of spatiotemporal micromobility information. The code implementing the dataset processing pipeline described in this section is publicly available in the accompanying GitHub repository \cite{Sahnoon2026AustinMicromobility}.

\subsection{Raw Dataset and Data Source}
\begin{figure}[hbt!]
    \centering
    \includegraphics[width=1\columnwidth]{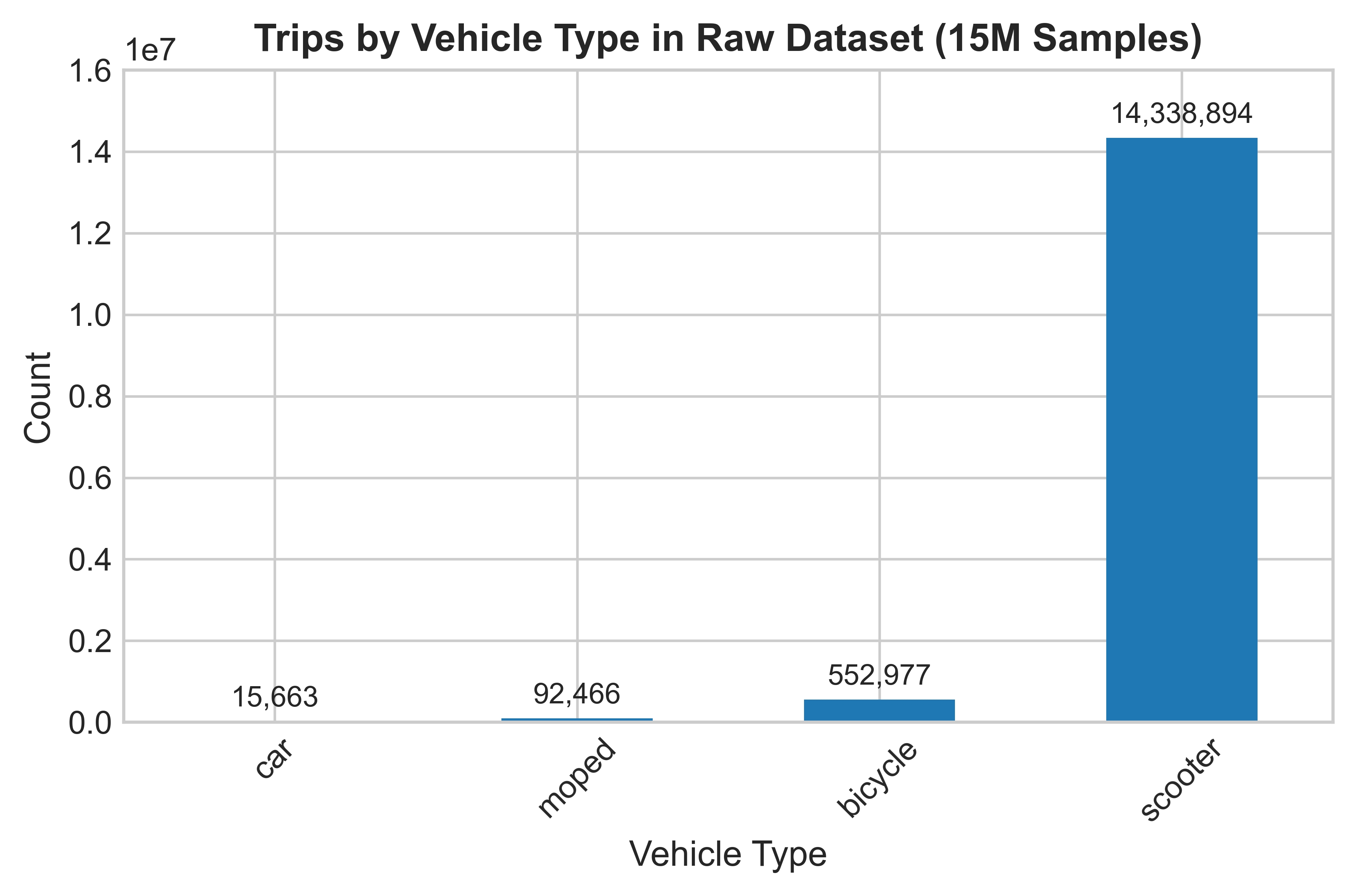}
    \caption{Trip counts per vehicle type in Austin's raw micromobility dataset (15 million samples).}
    \label{fig:vehicle_type_counts}
\end{figure}

\noindent The Shared Micromobility Vehicle Trips dataset is published by the City of Austin through its Open Data Portal \cite{Portaln.d.SharedMicromobility}. It spans the years 2018 to 2022 and contains approximately 15 million trip records. Each record includes 18 features describing temporal information, device identifiers, trip duration and distance, and the GEOID of the 2010 U.S. Census Tract associated with the trip origin and destination. As shown in Fig. \ref{fig:vehicle_type_counts}, the raw dataset covers different counts of trips for four shared micromobility modes: bicycles, car share vehicles, mopeds, and e-scooters. 

Exploratory data analysis of the full multi-year dataset revealed that trip records outside 2019 and non-e-scooter modes exhibit substantial temporal interruptions, including missing days and incomplete hourly coverage. Because this study requires chronologically consistent historical demand observations for lag identification and validation, the analysis is restricted to dockless e-scooter trips from the 2019 calendar year, which provide uninterrupted hourly coverage except for the missing hour between 2:00 a.m. and 2:59 a.m. on March 10, 2019, due to the spring Daylight Saving Time transition.

\subsection{Spatiotemporal Processing for Final Dataset Creation}

\noindent A multi-stage filtering and spatial processing procedure was applied to the raw 2019 e-scooter subset to retain valid trips with reliable temporal and spatial information, as summarized in Table \ref{table:filtering_criteria}. To assign spatial coordinates, the last six digits of each GEOID were matched to the corresponding 2010 Census Tract polygon from the Texas TIGER Line shapefiles \cite{Bureaun.d.2010TIGER}, and the tract centroid was used as the representative trip origin or destination. Trips with unmatched or invalid GEOIDs were removed. Trips were then restricted to tracts within the Austin municipal boundary using the official Austin jurisdictions shapefile \cite{Portaln.d.BOUNDARIESJurisdictions}, and trips associated with tracts outside the city limits were excluded. 

\begin{figure}[hbt!]
    \centering
    \includegraphics[width=1\columnwidth]{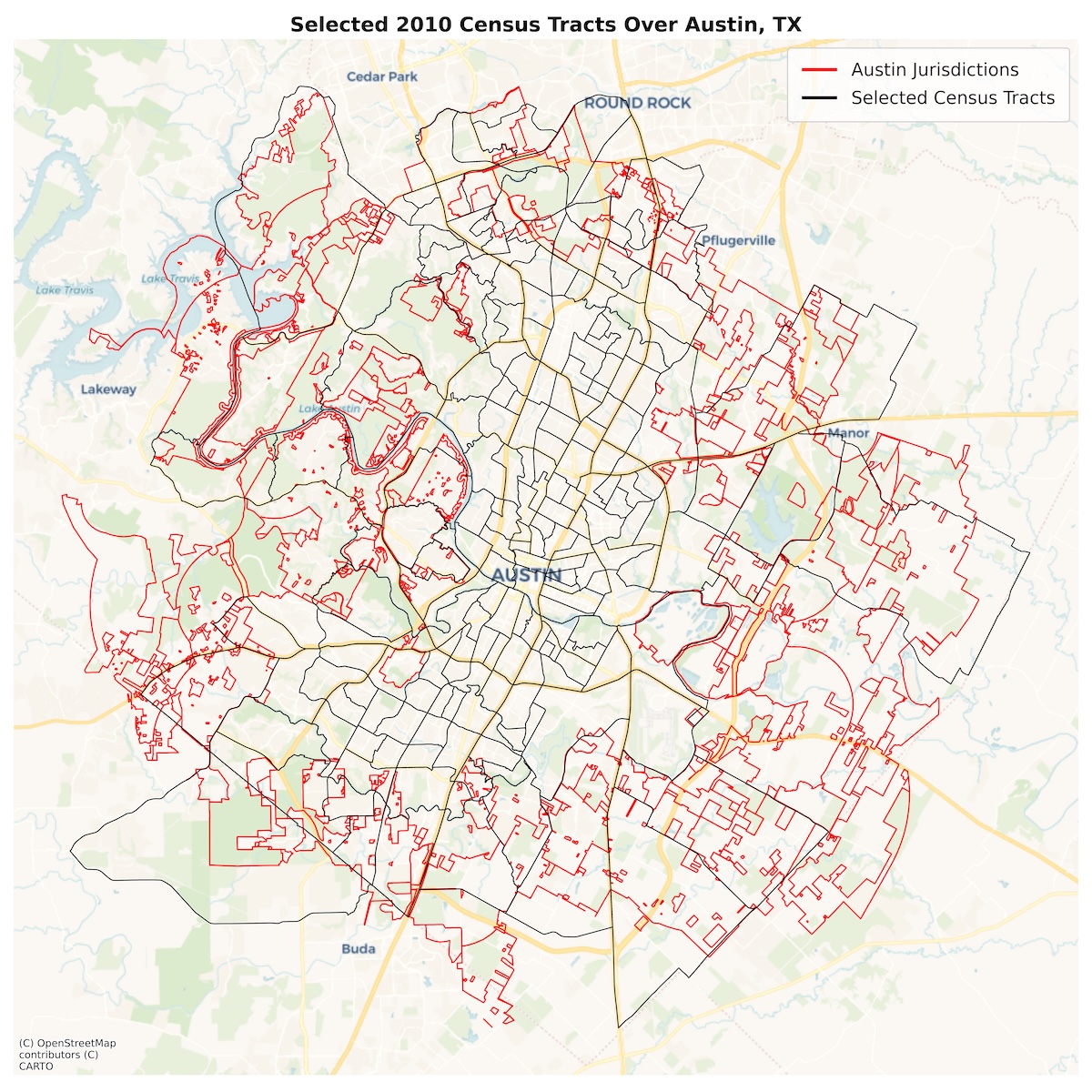}
    \caption{Spatial coverage of the final 2019 e-scooter dataset. The figure displays all 205 unique census tracts (in black) that appear as trip origins or destinations, along with the Austin city limits (in red) used for spatial filtering.}
    \label{fig:selected_tracts_map}
\end{figure}

After applying all filtering and spatial processing steps, the final dataset contains 4,939,008 valid e-scooter trips, representing approximately 86.6\% of the original 5,704,426 trips in the raw 2019 subset. These trips were performed by 52,907 unique e-scooter devices and span 199 distinct origin census tracts and 200 distinct destination census tracts, yielding a total of 205 unique tracts observed across all trip origins and destinations. Fig. \ref{fig:selected_tracts_map} illustrates the spatial distribution of these tracts within the city limits of Austin, TX, and highlights the final set of geographic areas represented in the dataset.

\begin{table}[t!]
\centering
\caption{Filtering criteria applied to construct the final Austin, TX 2019 e-scooter dataset.}
\label{table:filtering_criteria}
\begin{tabular}{p{0.28\linewidth} | p{0.60\linewidth}}
\hline
\textbf{Filter Category} & \textbf{Condition Applied} \\ 
\hline 
\hline

Vehicle type & Retain e-scooter trips only. \\[3pt]

Trip year & Retain trips occurring within the 2019 calendar year. \\[3pt]

Trip duration & Keep trips between 1 and 120 minutes inclusive. \\[3pt]

Trip distance & Keep trips between 0.1 and 35 kilometers inclusive. \\[3pt]

Average speed & Keep trips with speed between 2 and 26 kilometers per hour inclusive. \\[3pt]

Tract GEOID & Trip origin and destination must have valid census tract identifiers. \\[3pt]

Spatial coordinates & Trips are retained only if valid centroid coordinates can be assigned to both the origin and destination census tracts within the city limits. \\ \hline

\end{tabular}
\end{table}

\subsection{Final Dataset Statistics and Spatiotemporal Demand Patterns}

\begin{figure*}[!t]
\centering
    \begin{minipage}{0.32\textwidth}
        \centering
        \includegraphics[width=\linewidth]{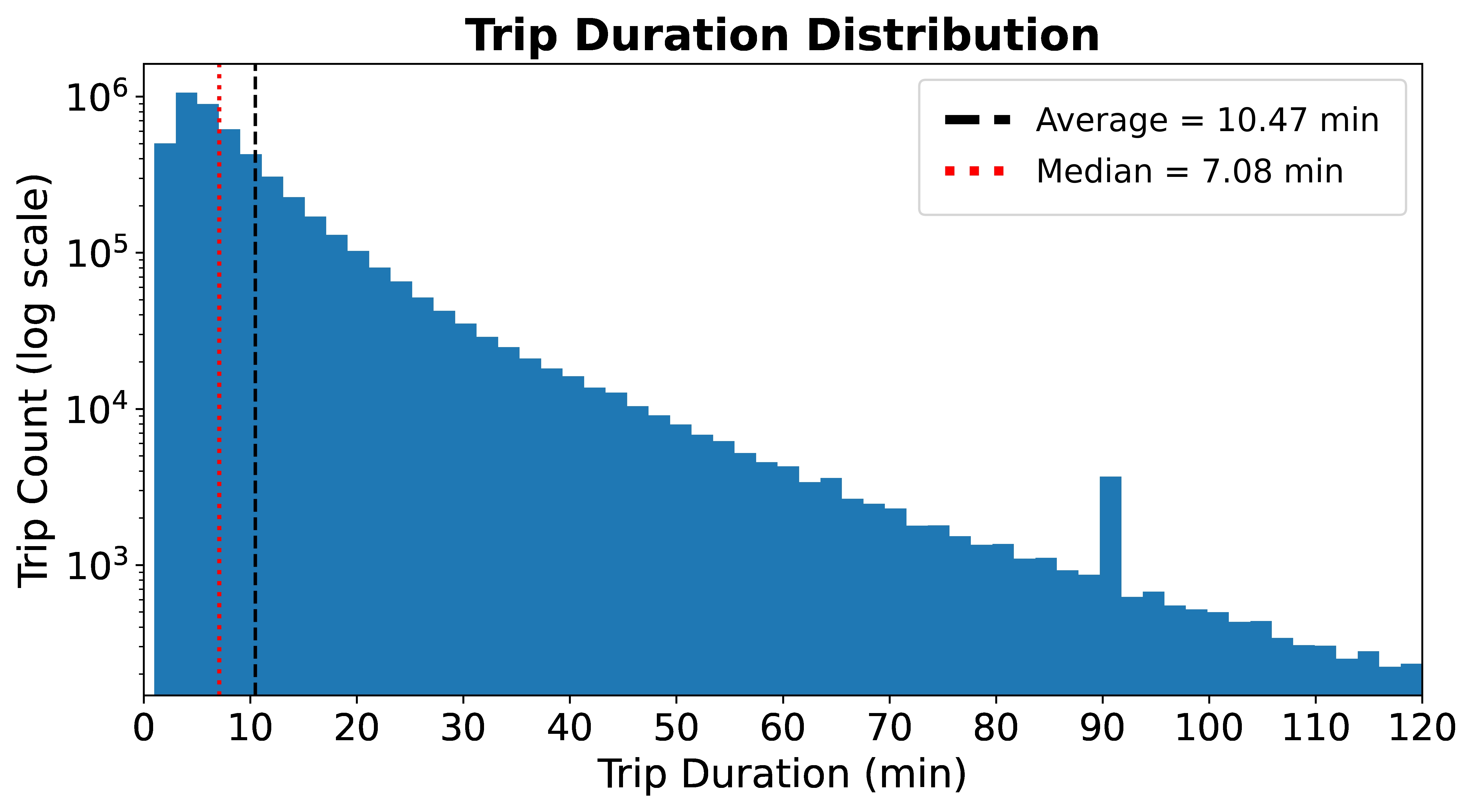}
        {\scriptsize (a) Trip duration distribution in minutes.\par}
    \end{minipage}
    \hfill
    \begin{minipage}{0.32\textwidth}
        \centering
        \includegraphics[width=\linewidth]{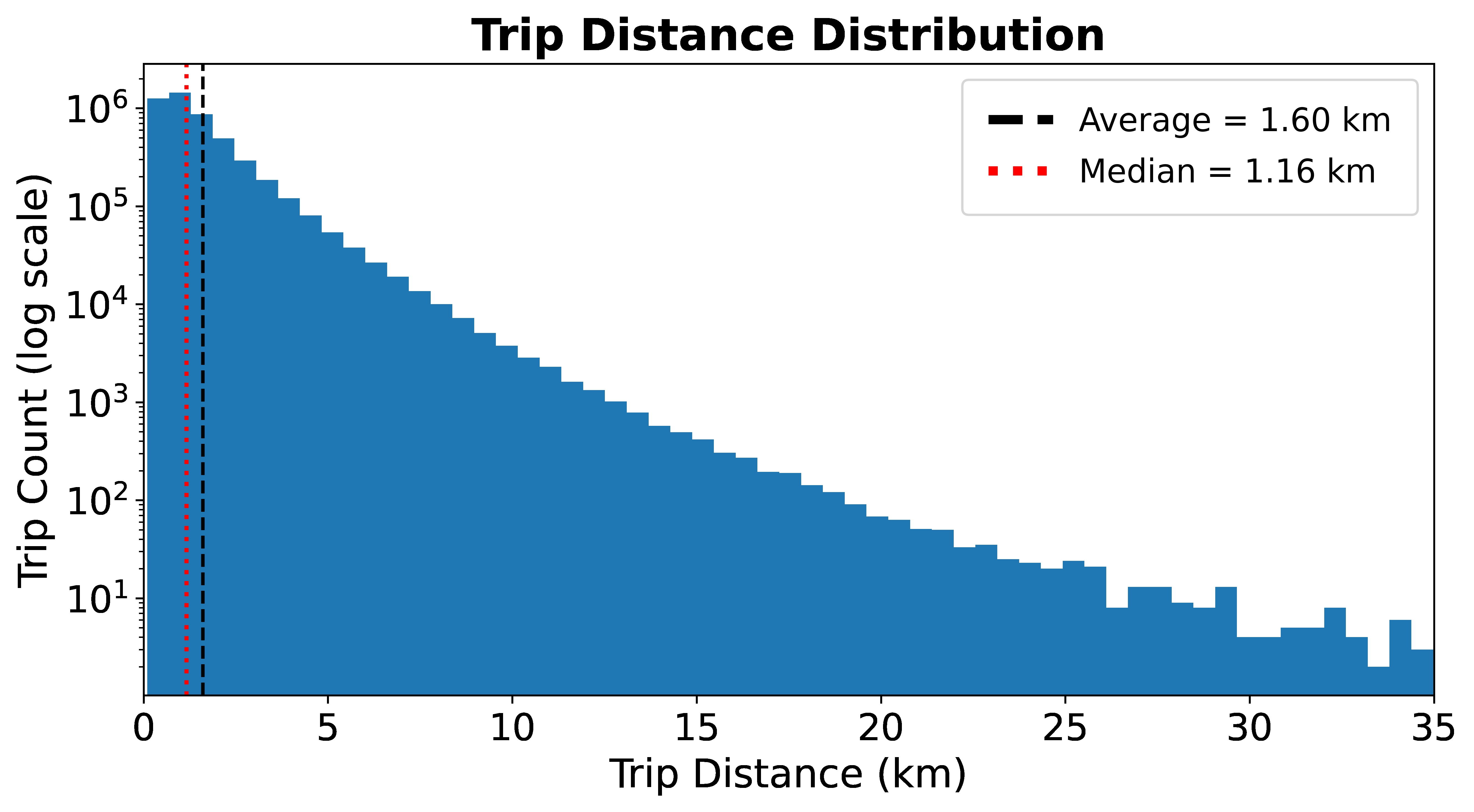}
        {\scriptsize (b) Trip distance distribution in kilometers.\par}
    \end{minipage}
    \hfill
    \begin{minipage}{0.32\textwidth}
        \centering
        \includegraphics[width=\linewidth]{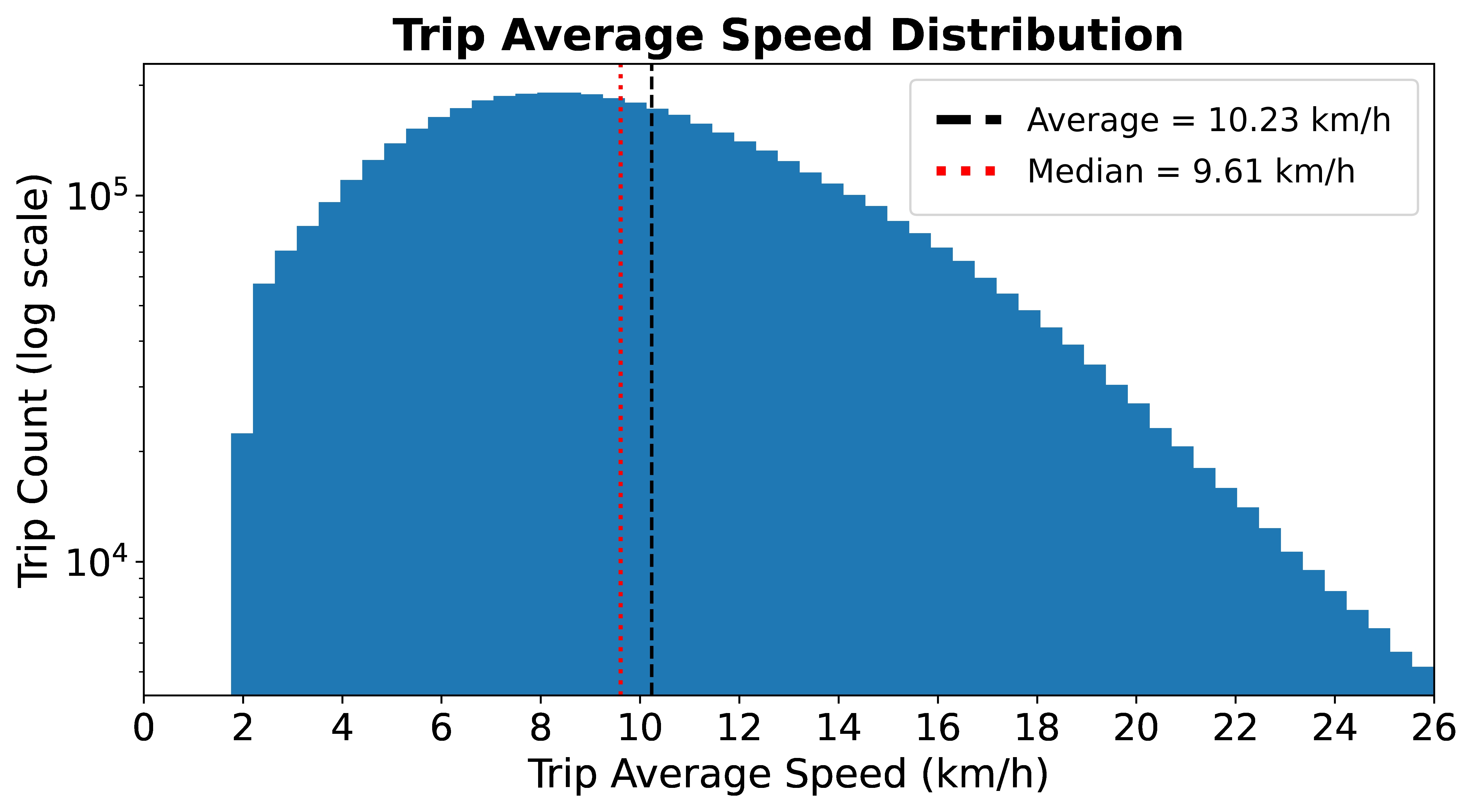}
        {\scriptsize (c) Trip average speed distribution in kilometers per hour.\par}
    \end{minipage}
    \caption{Distribution of core trip characteristics for the processed 2019 e-scooter dataset: (a) trip duration in minutes, (b) trip distance in kilometers, and (c) trip average speed in kilometers per hour.}

    \label{fig:trip_distributions}
\end{figure*}

\noindent The filtered e-scooter dataset reflects short distance urban trips typical of shared micromobility services. On average, trips lasted 10.47 minutes and covered 1.60 kilometers, with median values of 7.08 minutes and 1.16 kilometers respectively. The average trip speed is 10.23 kilometers per hour with a median of 9.61 kilometers per hour. These trip distributions are shown in Fig. \ref{fig:trip_distributions}.

\begin{figure}[hbt!]
    \centering
    \includegraphics[width=1\columnwidth]{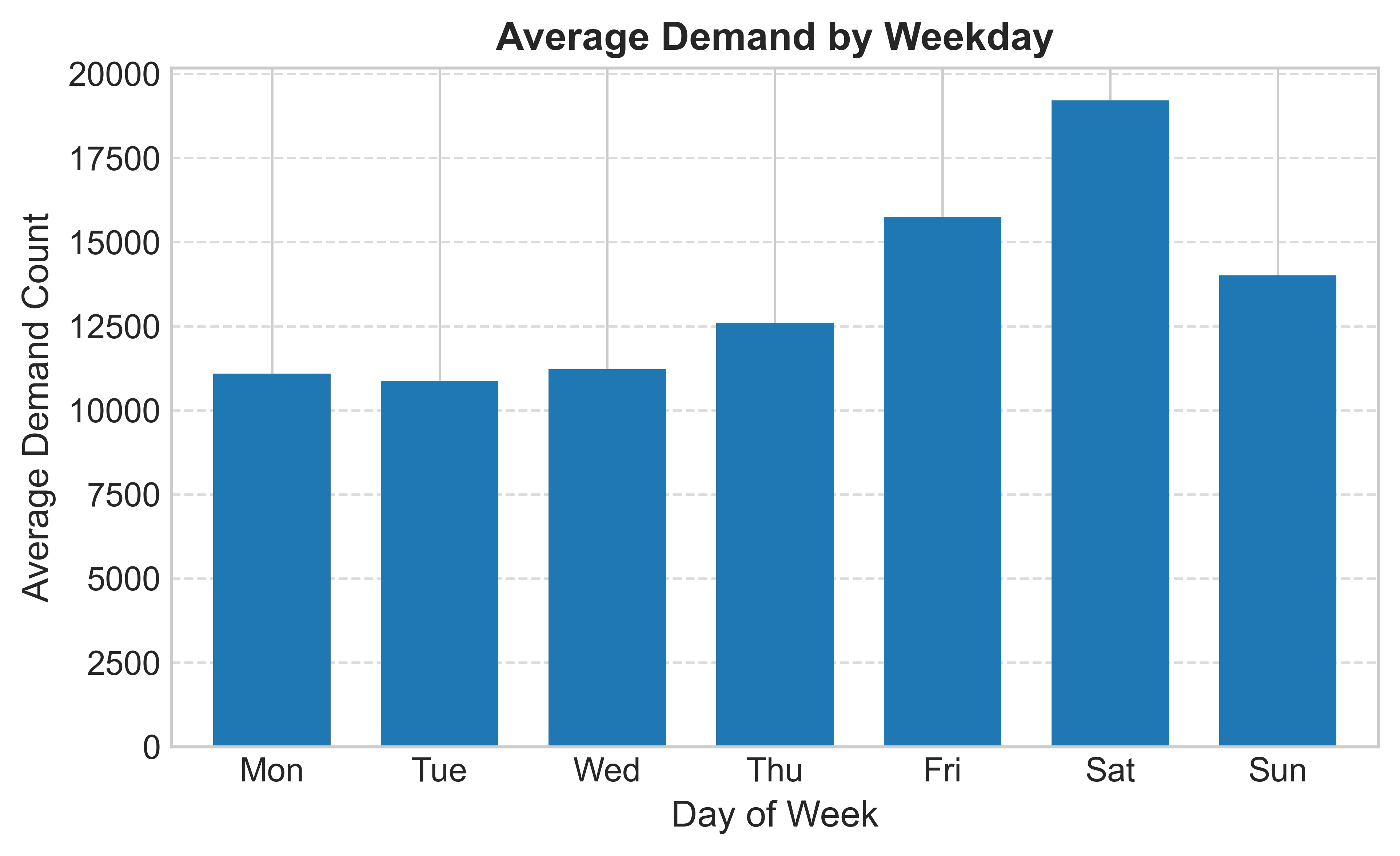}
    \caption{Average daily e-scooter demand in 2019, obtained by averaging daily trip counts across all days of the same weekday.}
    \label{fig:daily_demand}
\end{figure}

\begin{figure}[hbt!]
    \centering
    \includegraphics[width=1\columnwidth]{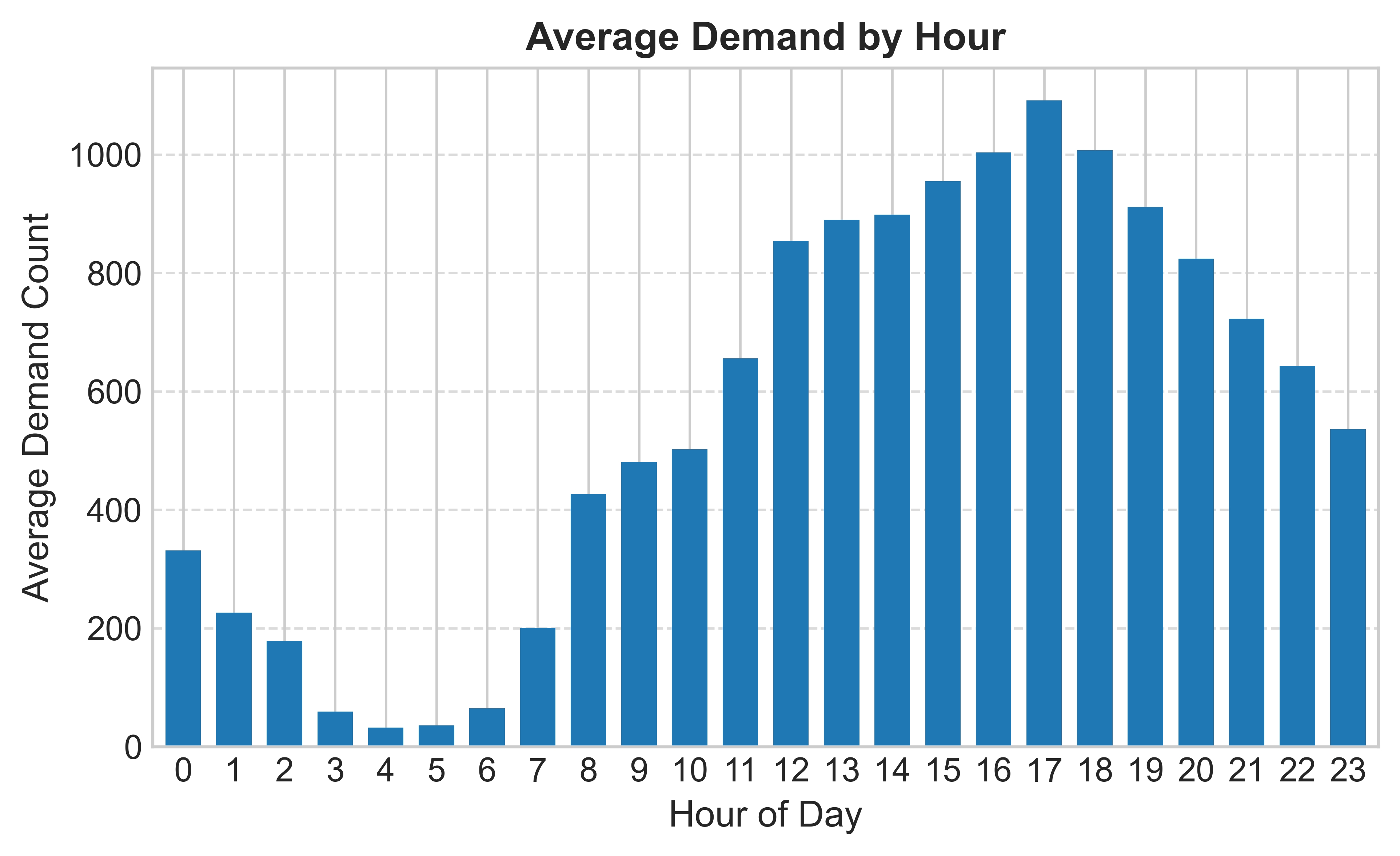}
    \caption{Average hourly e-scooter demand in 2019, calculated by averaging trip counts for each hour of the day across all calendar days.}
    \label{fig:hourly_demand}
\end{figure}

To characterize the temporal structure of e-scooter demand, average daily and hourly usage was computed over the full processed 2019 dataset. Fig. \ref{fig:daily_demand} shows that demand is relatively stable from Monday through Thursday, rises on Friday, peaks on Saturday, and remains elevated on Sunday, indicating strong weekend recreational and leisure use. Fig. \ref{fig:hourly_demand} shows that the hourly demand is lowest overnight, increases from about 6:00 a.m., peaks between approximately 3:00 p.m. and 6:00 p.m., and then declines into the late evening. Unlike traditional traffic peaks associated with work commuting, this pattern reflects more flexible, short-distance, and recreational urban mobility.

\begin{figure*}[!t]
\centering
    \begin{minipage}{0.45\textwidth}
        \centering
        \includegraphics[width=\linewidth]{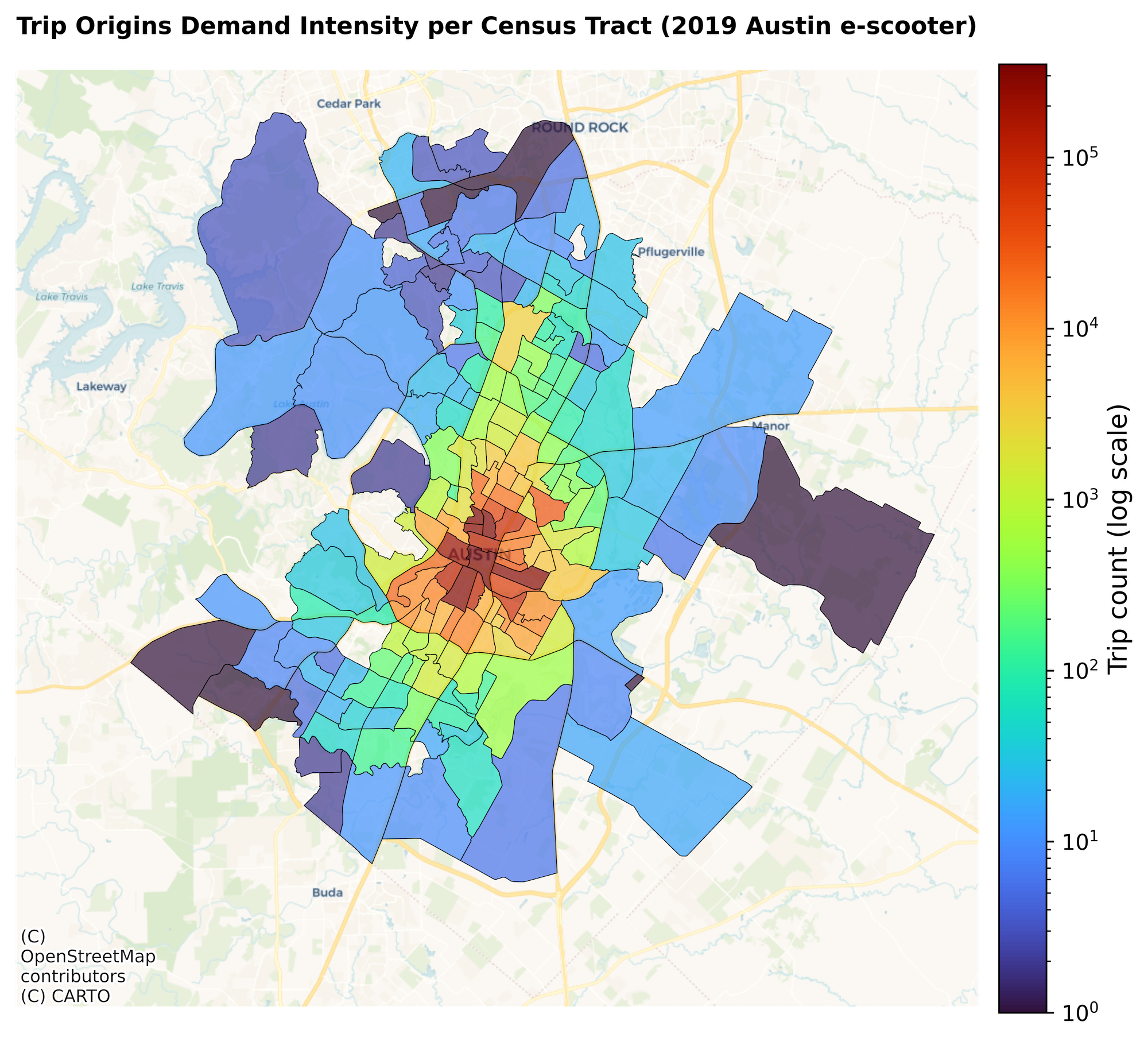}
        {\scriptsize (a) Trip origin demand intensity per census tract.\par}
    \end{minipage}
    \hfill
    \begin{minipage}{0.45\textwidth}
        \centering
        \includegraphics[width=\linewidth]{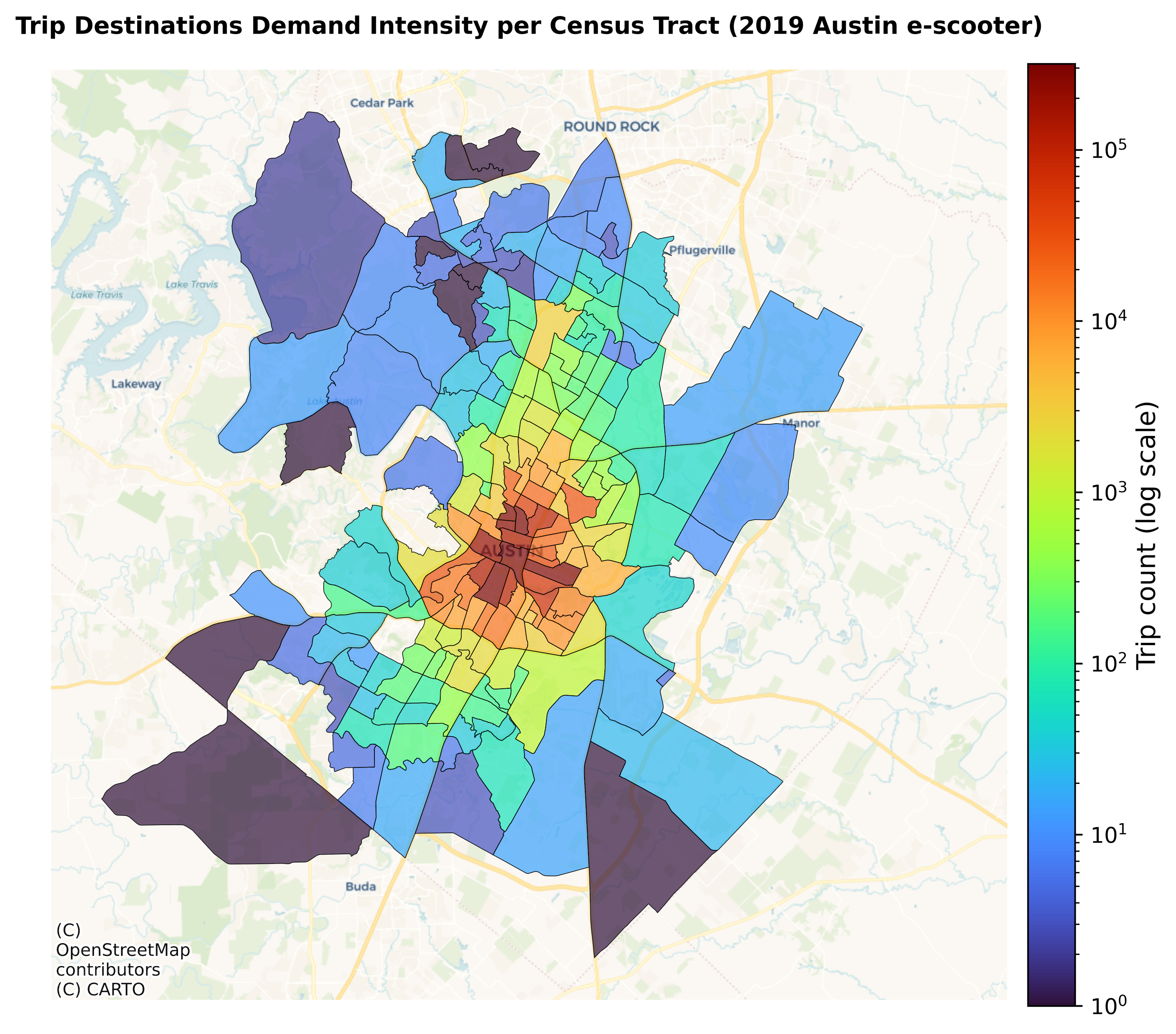}
        {\scriptsize (b) Trip destination demand intensity per census tract.\par}
    \end{minipage}
    \caption{Spatial distribution of (a) trip origin and (b) trip destination intensities for Austin's 2019 e-scooter dataset at the census tract level, highlighting high-demand tracts in central Austin, TX.}

    \label{fig:trip_spatial_distributions}
\end{figure*}

The spatial distribution of Austin's e-scooter activity during the 2019 calendar year was examined by aggregating all trip origins and destinations at the census tract level. Fig. \ref{fig:trip_spatial_distributions} shows the resulting pick-up and drop-off intensities across the city. The highest concentrations of demand are located in central Austin, particularly in and around Downtown, the University of Texas at Austin campus, and the surrounding mixed-use neighborhoods where population density, employment centers, and entertainment districts are clustered. Elevated activity is also observed along major commercial corridors such as South Congress, East Austin near the restaurant and nightlife districts, and the North Lamar–Guadalupe corridor connecting the campus area to central residential zones. In contrast, tracts near the city edges exhibit substantially lower pick-up and drop-off volumes. These patterns are consistent with higher urban activity, multimodal accessibility, and concentrations of trip-attracting destinations in the city center.

\subsection{Transformation of Trip Records into Grid-Based Hourly Demand Images}
\label{sec:image-gen}
\noindent To support deep learning models that operate on spatial grids, all valid trips were transformed into hourly images representing the distribution of pick-up and drop-off activity across the city. This required projecting the tract centroids from the World Geodetic System 1984 (WGS84) into the Universal Transverse Mercator coordinate system, UTM Zone 14 North. UTM expresses point locations in meters, which enables the construction of a uniform grid for rasterization.

\begin{figure}[hbt!]
    \centering
    \includegraphics[width=1\columnwidth]{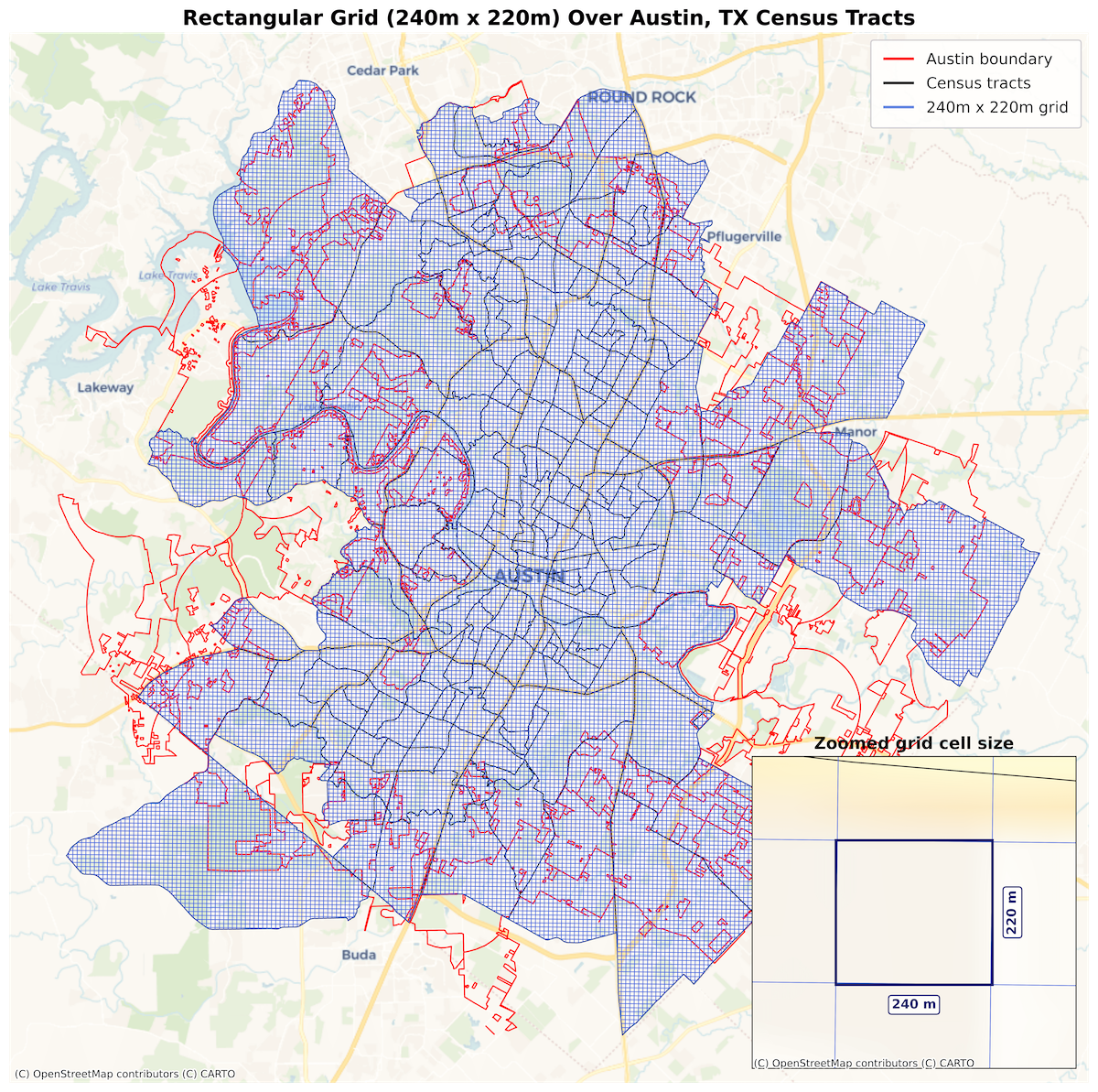}
    \caption{Generation of the $240\,m\,\times\,220\,m$ rectangular grid over the selected census tracts within the Austin, TX municipal boundary.}
    \label{fig:rectangular_grid}
\end{figure}

A rectangular grid covering the selected census tracts within the Austin, TX municipal boundary was constructed using fixed cell dimensions of $240\,m\,\times\,220\,m$. Fig. \ref{fig:rectangular_grid} shows the spatial extent of the grid and its alignment with the study area. This spatial resolution reflects realistic access distances for e-scooter users, assuming an average walking speed of 1.47 meters per second \cite{Bowman1994PEDESTRIANWALKING}. Under this assumption, the grid cell diagonal of approximately 325 meters corresponds to a worst-case walking time of about 4 minutes. The resulting representation is intentionally fine-grained, but because trip origins and destinations are reported at the census tract level rather than as precise coordinates, many cells remain sparse after rasterization. The chosen resolution nevertheless preserves localized spatial structure, while sparsity is addressed later through the masking and modelling framework.

Let \( r \) and \( c \) denote the row and column indices of the grid. Let \( d \) represent a calendar day, \( h \) represent the hour of day, and \( (x_{i}^{s}, y_{i}^{s}) \) and \( (x_{i}^{e}, y_{i}^{e}) \) denote the start and end UTM coordinates of trip \( i \), respectively. For each pair \( (d, h) \), two images were generated. The pick-up image counts all trips starting in hour \( h \) of day \( d \), and the drop-off image counts all trips that end during the same day and hour. The pixel values for the pick-up image \( I_{p}(r,c,d,h) \) and drop-off image \( I_{d}(r,c,d,h) \) were computed as:

\begin{equation}
\label{eq:pickup-img}
\begin{aligned}[b]
I_{p}(r,c,d,h) = \sum_{i \in \mathcal{T}_{\text{start}}(d,h)} \mathbf{1}\{ (x_i^{s},y_i^{s}) \in \text{cell}(r,c) \}
\end{aligned}
\end{equation}

\begin{equation}
\label{eq:dropoff-img}
\begin{aligned}[b]
I_{d}(r,c,d,h) = \sum_{i \in \mathcal{T}_{\text{end}}(d,h)} \mathbf{1}\{ (x_i^{e},y_i^{e}) \in \text{cell}(r,c) \}
\end{aligned}
\end{equation}

\noindent where \( \mathcal{T}_{\text{start}}(d,h) \) is the set of trips with start day \( d \) and hour \( h \) and \( \mathcal{T}_{\text{end}}(d,h) \) is the set of trips with end times on the same day and hour. The indicator function \( \mathbf{1}\{\cdot\} \) equals one when the coordinate lies within the grid cell and zero otherwise. 

\begin{figure*}[hbt!]
\centering
    \begin{minipage}{0.32\textwidth}
        \centering
        \includegraphics[width=\linewidth]{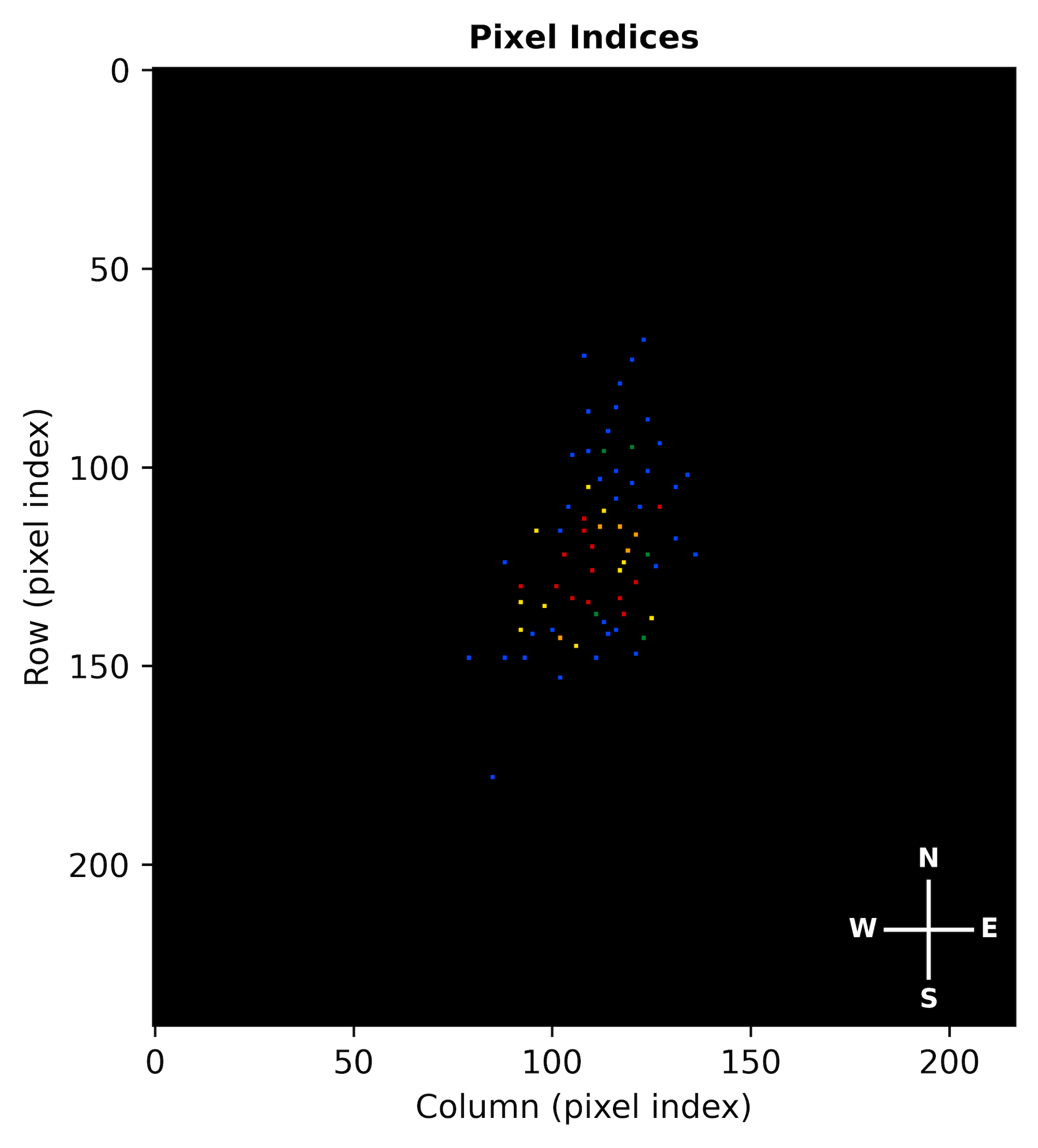}
        {\scriptsize (a) Pick-up sample image with pixel indices coordinates.\par}
    \end{minipage}
    \hfill
    \begin{minipage}{0.32\textwidth}
        \centering
        \includegraphics[width=\linewidth]{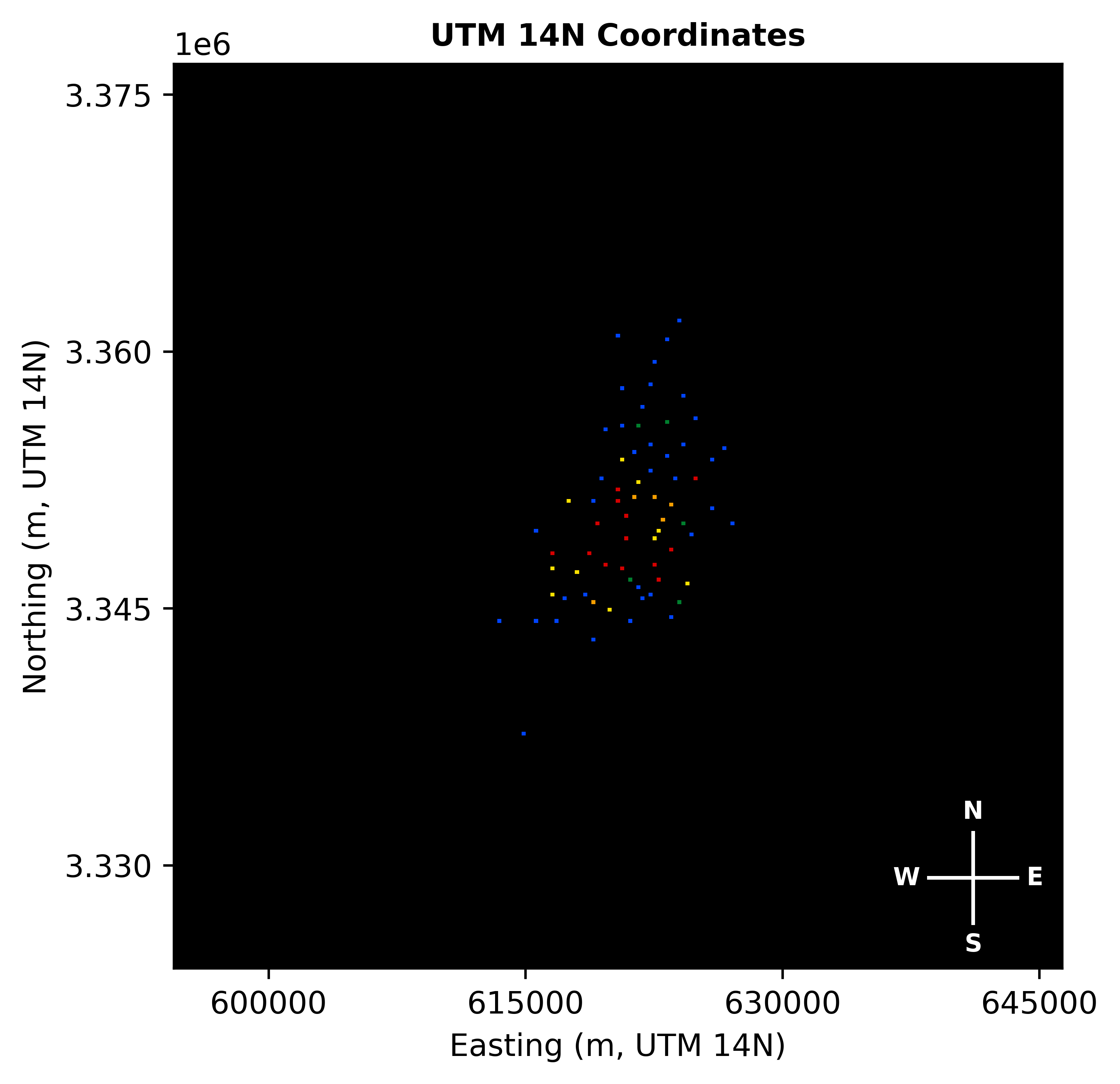}
        {\scriptsize (b) Pick-up sample image with UTM Zone 14N coordinates.\par}
    \end{minipage}
    \hfill
    \begin{minipage}{0.32\textwidth}
        \centering
        \includegraphics[width=\linewidth]{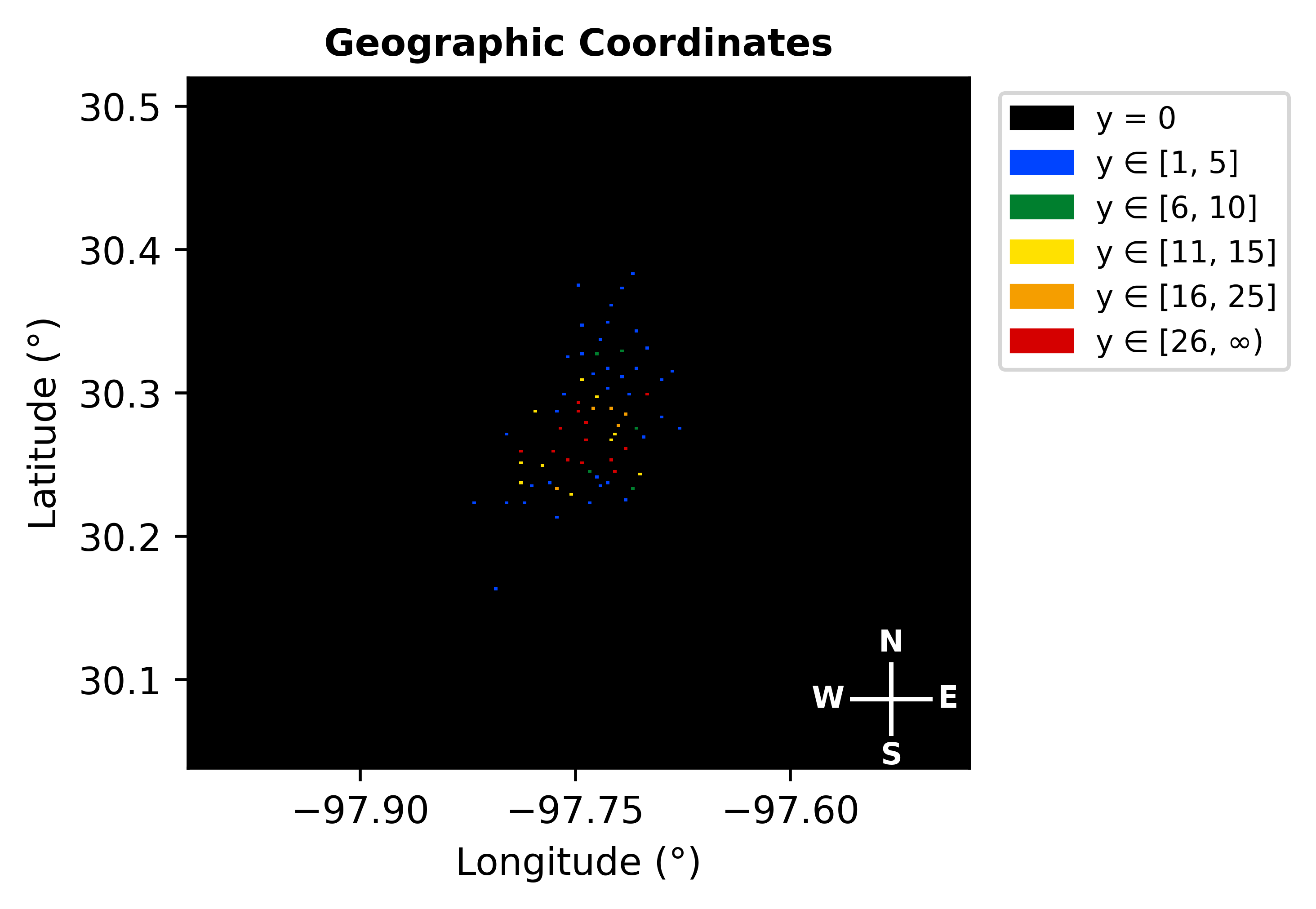}
        {\scriptsize (c) Pick-up sample image with geographic coordinates.\par}
    \end{minipage}
    \caption{Sample pick-up demand image generated for 4:00 p.m. on March 17, 2019. Subplot (a) uses pixel-based indexing of the raster grid, subplot (b) expresses the same demand field in projected UTM Zone 14N coordinates, and subplot (c) shows the demand field in geographic coordinates. The color legend indicates the ground-truth demand level associated with each pixel value and is included for visualization only. For model input, demand images are represented in black and white, where the intensity of each white pixel encodes the demand level.}
    \label{fig:pickup_demand_img_samples}
\end{figure*}

The resulting image resolution is 241\,\texttimes\,217 pixels, where 241 is the height and 217 is the width. All images were stored as 16-bit Portable Network Graphics (PNG) files to preserve exact demand counts. In total, 17,520 images were generated: 8,760 hourly pick-up images and 8,760 hourly drop-off images. Fig. \ref{fig:pickup_demand_img_samples} presents a representative sample image of pick-up demand, illustrating the rasterized demand field using pixel indices, projected UTM Zone 14N coordinates, and geographic coordinates. Similar spatial patterns are observed for drop-off demand, confirming consistent spatial structure and coordinate alignment across both demand types.

\subsection{Construction of the Global Binary Activity Mask}
\label{subsec:global-mask}

\noindent A global binary activity mask is constructed to identify all spatial grid cells where e-scooter activity occurs at least once over the temporal span of the dataset. This mask defines the spatial support over which model training and evaluation are performed, ensuring that error metrics are computed only within the historically active region.

Let \( M_{rc} \) denote the global binary activity mask value at grid cell \( (r,c) \). The global binary activity mask is defined as:
\begin{equation}
\label{eq:binary-mask}
M_{rc} =
\begin{cases}
1, & \text{if } \sum_{t} \big( p(r,c,t) + d(r,c,t) \big) > 0 \\
0, & \text{otherwise}
\end{cases}
\end{equation}

\noindent where \( p(r,c,t) \) and \( d(r,c,t) \) denote the pick-up and drop-off demand at grid cell \( (r,c) \) and time \( t \), respectively. A grid cell is marked as active if any pick-up or drop-off demand has historically occurred at that location; otherwise, it is marked as inactive. Therefore, the set of spatial grid cells corresponding to historically active demand locations is defined as:

\begin{equation}
\label{eq:Omega}
\Omega = \{ (r,c) \mid M_{rc} = 1 \}
\end{equation}

\begin{figure*}[hbt!]
\centering
    \begin{minipage}{0.32\textwidth}
        \centering
        \includegraphics[width=\linewidth]{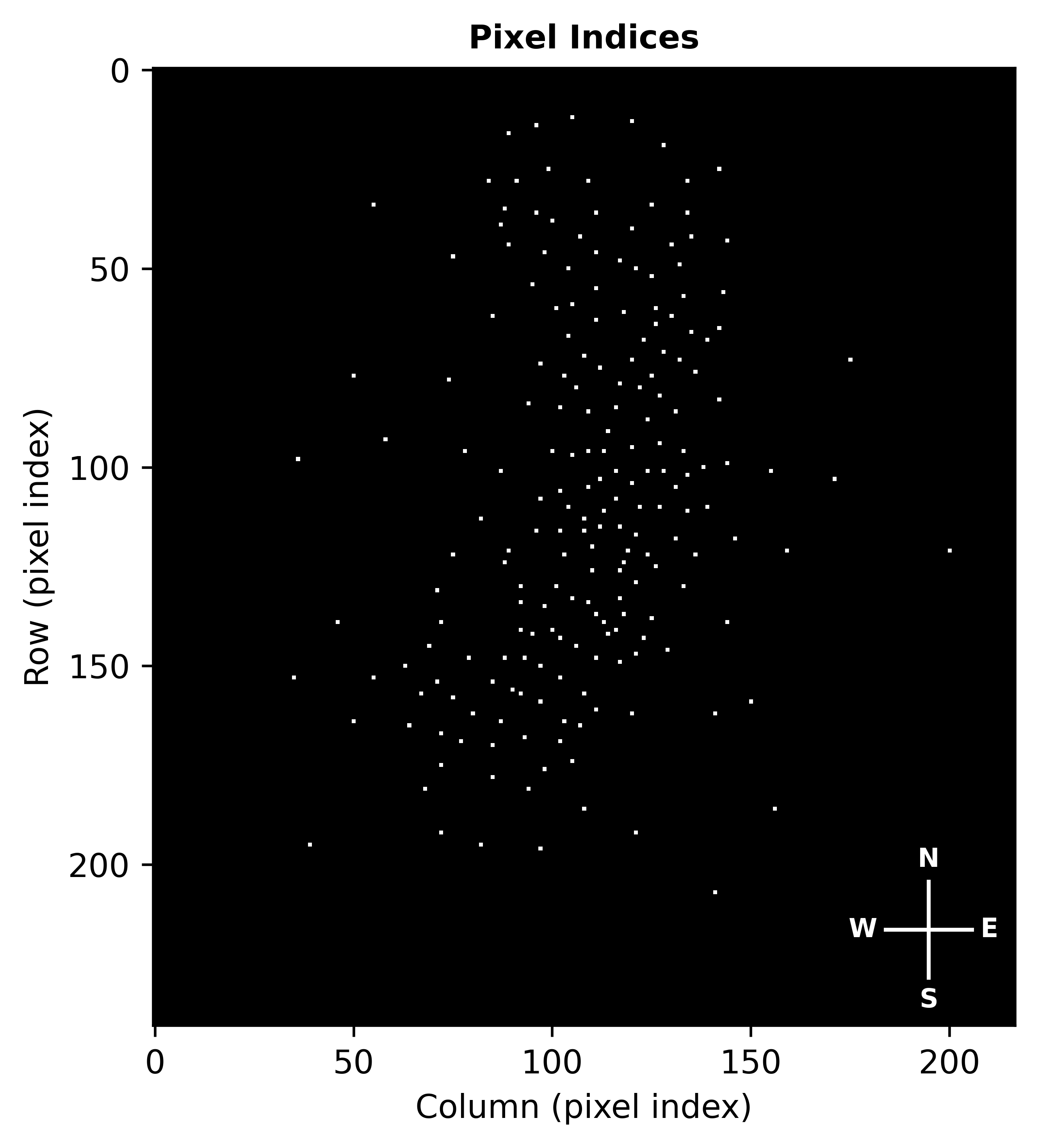}
        {\scriptsize (a) Global binary activity mask with pixel indices coordinates.\par}
    \end{minipage}
    \hfill
    \begin{minipage}{0.32\textwidth}
        \centering
        \includegraphics[width=\linewidth]{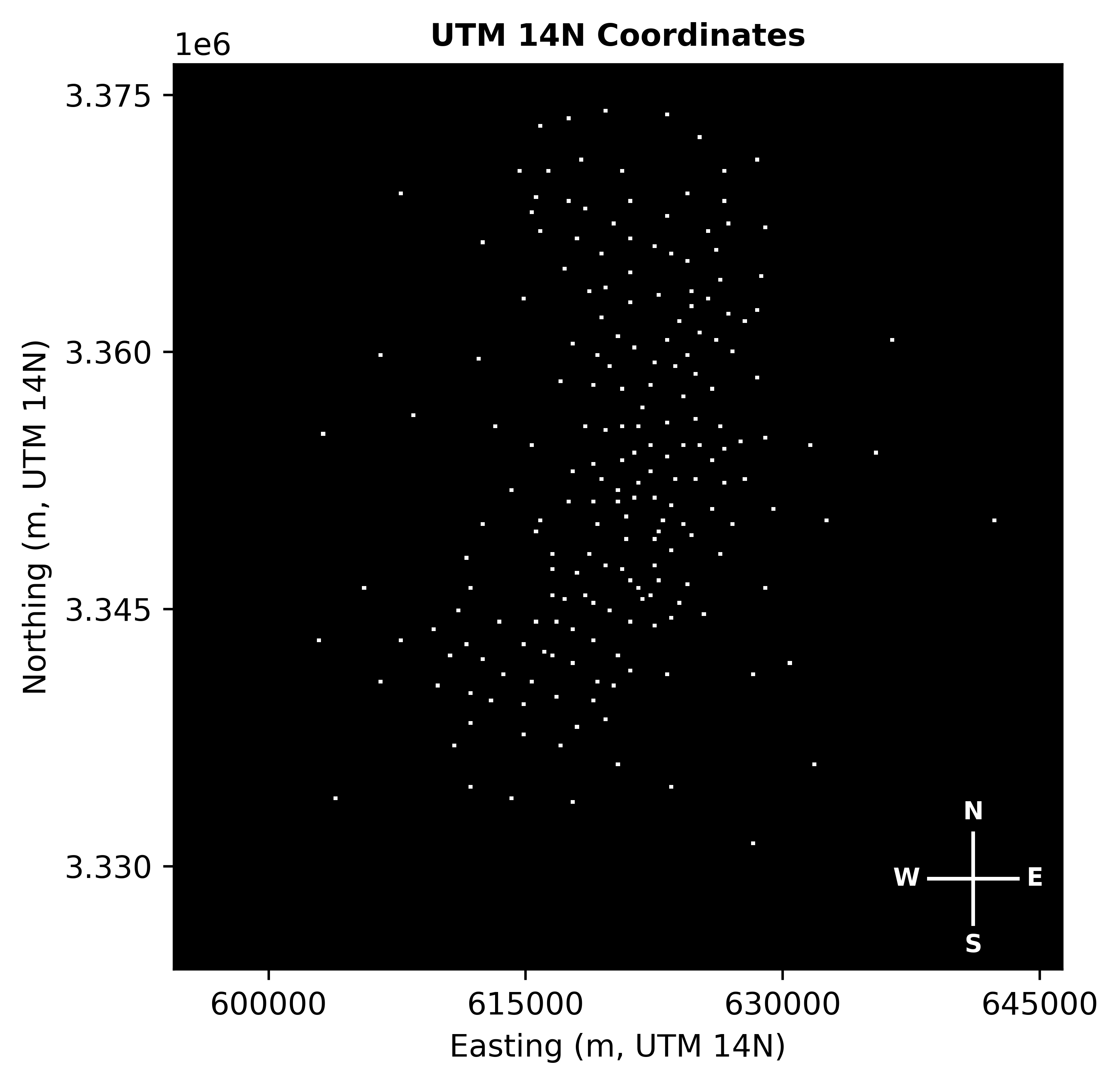}
        {\scriptsize (b) Global binary activity mask with UTM Zone 14N coordinates.\par}
    \end{minipage}
    \hfill
    \begin{minipage}{0.32\textwidth}
        \centering
        \includegraphics[width=\linewidth]{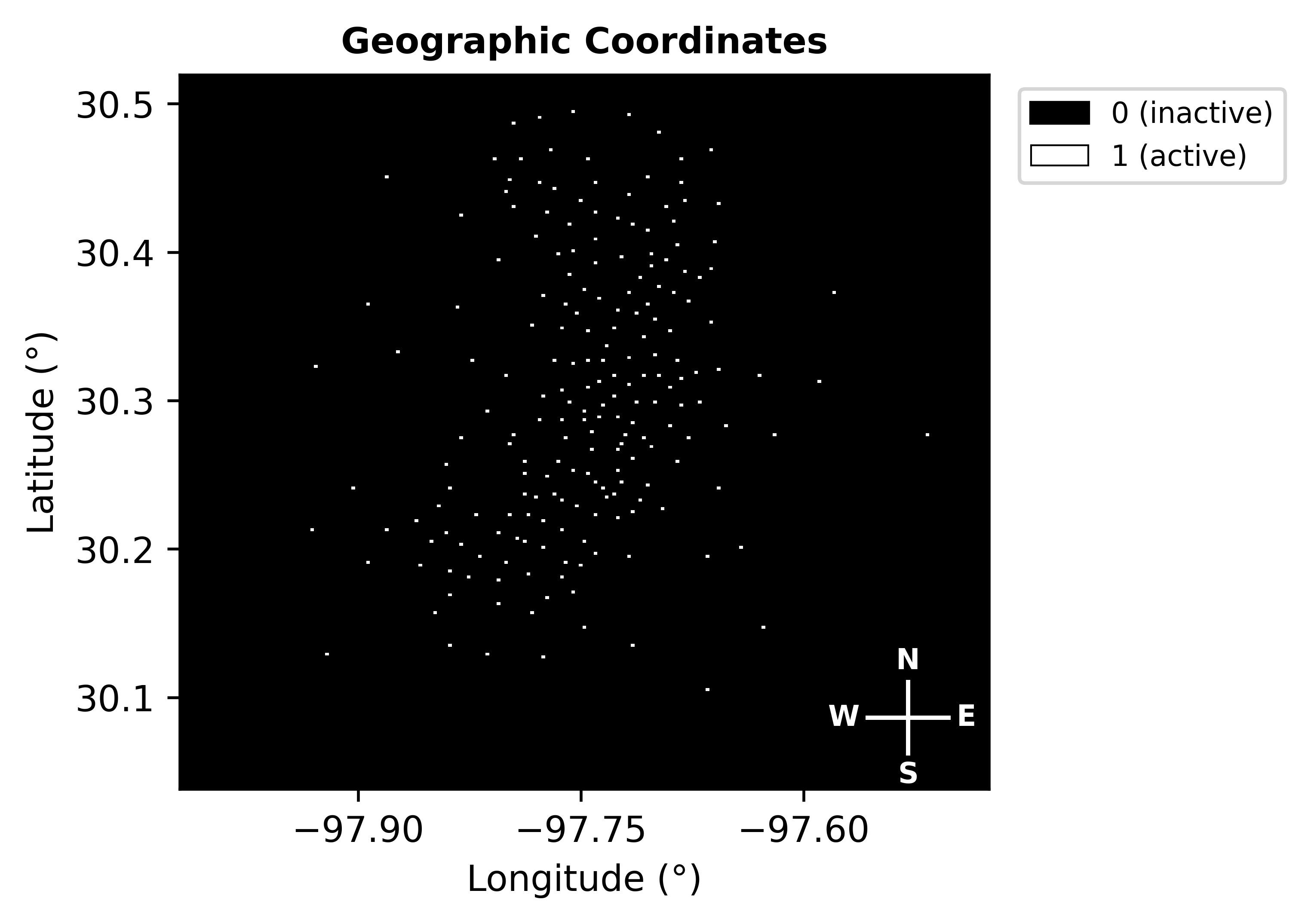}
        {\scriptsize (c) Global binary activity mask with geographic coordinates.\par}
    \end{minipage}
    \caption{Global binary activity mask used to define the operational footprint of the historically active e-scooter demand locations to guide model training and evaluation across both the pick-up and drop-off channels. Subplot (a) uses pixel-based indexing of the raster grid, subplot (b) expresses the same demand field in projected UTM Zone 14N coordinates, and subplot (c) shows the demand field in geographic coordinates.}
    \label{fig:global_mask}
\end{figure*}

All loss and evaluation metrics reported in this study, including the training loss and error-based metrics as well as the coefficient of determination ($R^2$), are computed exclusively over the active spatial support set \( \Omega \). The global binary activity mask captures the operational footprint of the shared e-scooter system and provides spatial consistency across all deep learning experiments. The same mask is applied uniformly to both pick-up and drop-off demand prediction tasks. A visualization of the global binary activity mask using pixel indices, projected UTM Zone 14N coordinates, and geographic coordinates appears in Fig. \ref{fig:global_mask}.

\section{Experimental Setup}
\label{sec:exp-design}
\noindent This section describes the experimental protocol used to evaluate temporal input configurations for short-term e-scooter demand prediction. We define the prediction horizons and experimental scope, construct leakage-free train, validation, and test subsets, and specify the deep learning input representation used throughout the study. We then summarize the baseline model architecture, loss function formulation, evaluation metrics, and statistical testing procedures used to support rigorous comparisons across temporal input configurations.

\subsection{Prediction Horizons and Experimental Scope}
\noindent This study addresses two related prediction tasks: next-hour demand prediction and next 24-hour demand prediction. Next-hour prediction supports short-term operational decision-making (e.g., dynamic rebalancing and resource allocation), while next 24-hour prediction provides value for system-level planning and daily deployment strategies. Together, these two horizons enhance the practical usefulness of the modelling framework across multiple temporal scales. Both tasks aim to estimate spatial pick-up and drop-off demand maps (i.e., images), but differ in their temporal prediction horizon. 

To ensure consistency and comparability, the same temporal dataset split, spatial representation, and preprocessing pipeline are used for both prediction horizons. All subsequent methodological steps related to the construction of spatiotemporal input tensors, selection of historical time steps, and determination of the optimal number of input channels follow the same procedure but are applied independently for each task. In both cases, each sample consists of a target demand image paired with a sequence of historical demand images extending backward in time, with the only distinction between the two tasks being the temporal configuration of the input lags, which determines both the selection of temporal lags at each channel and how far back in time the historical input extends.

\subsection{Temporal Dataset Splitting and Leakage Prevention}
\label{subsec:dataset_splitting}
\noindent The dataset is partitioned into training, validation, and test subsets using strictly non-overlapping temporal windows to prevent any form of data leakage. The same temporal split is applied consistently across all experiments and prediction horizons. At this stage, a conservative maximum lookback window of 504 hours, equivalent to 21 days, is assumed for all samples to ensure that the historical demand patterns used for each target are drawn from a period of less than one month, which minimizes the influence of seasonal shifts and long-term demand variations that are outside the scope of short-term demand prediction. While this 504-hour window defines the maximum allowable lookback for splitting purposes, the actual temporal extent of the model input is subsequently revisited and optimized for each prediction horizon (i.e., next-hour and next 24-hour) in Section \ref{optimal-depth}.

Each sample corresponds to a target hour for which demand is predicted and is initially associated with a sequence of historical demand images extending up to the assumed maximum lookback of 504 hours. To guarantee complete temporal separation between subsets, an additional one-hour buffer is enforced between consecutive splits. As a result, the first target hour of a given subset is always 505 hours away from the final target hour of the preceding subset. This design ensures that no historical input used by a sample in one subset can overlap with either the target or the historical input windows of another subset, eliminating any possibility of temporal information leakage.

Under this protocol, the Austin, TX, 2019 e-scooter dataset contains a total of 7,248 samples. The training subset comprises 3,743 samples, representing 51.64\%  of the full dataset, with target timestamps ranging from 12:00 a.m. on 22 January 2019 to 10:00 p.m. on 26 June 2019. The validation subset contains 1,752 samples, accounting for 24.17\% of the dataset, with target timestamps spanning from 11:00 p.m. on 17 July 2019 to 10:00 p.m. on 28 September 2019. A temporal gap of 505 hours is maintained between the end of the training subset and the start of the validation subset. The test subset includes 1,753 samples, corresponding to 24.18\% of the dataset, with target timestamps ranging from 11:00 p.m. on 19 October 2019 to 11:00 p.m. on 31 December 2019. As with the previous split, a 505 hour separation is enforced between the validation and test subsets.

\subsection{Input Representation and Spatiotemporal Channel Construction}
\label{subsec:Input_channels}

\noindent The model input is constructed to preserve both the spatial and temporal characteristics of e-scooter demand. Spatial structure is encoded through hourly pick-up and drop-off demand intensity images generated using the census-tract-to-grid transformation described in Section \ref{sec:image-gen}, where each image represents a fixed spatial grid of aggregated demand counts for a given hour.

\begin{figure*}[!t]
\centering
\includegraphics[width=\textwidth]{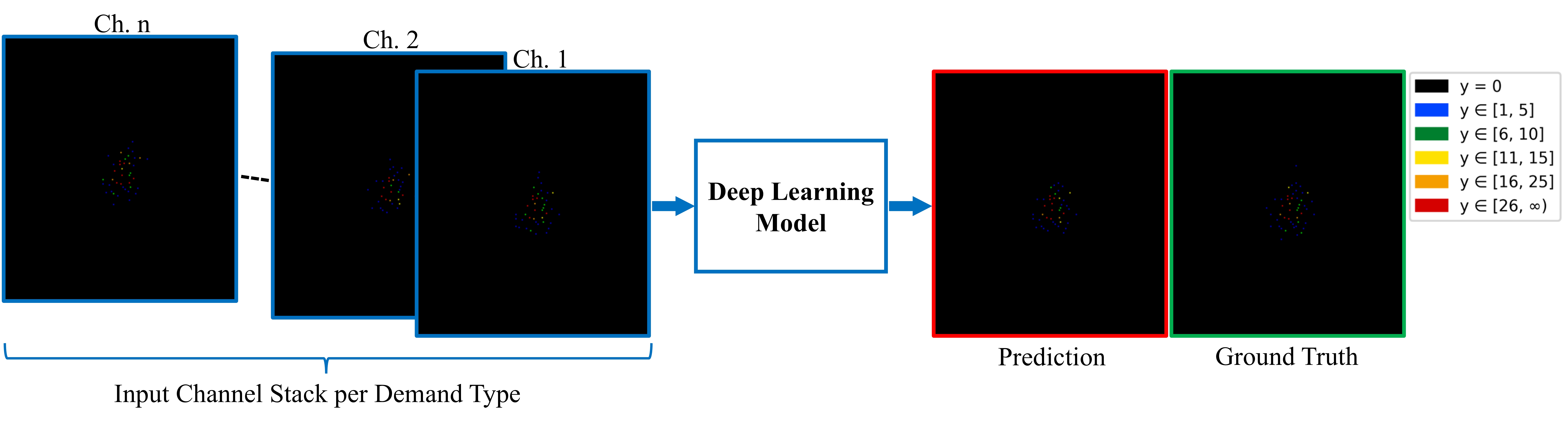}
\caption{Channel-based input representation for spatiotemporal demand prediction. The figure illustrates the pick-up demand case, where historical demand images are stacked as ordered channels (e.g., $\mathrm{Ch.}~1,\mathrm{Ch.}~2,\ldots,\mathrm{Ch.}~n$). The same construction is extended to drop-off demand by appending the corresponding channels and producing a two-channel output, which is evaluated against the ground-truth target map. Pixel color coding is for demand-level visual illustration only.}
\label{fig:input_representation_flowchart}
\end{figure*}

Temporal context is introduced by stacking historical demand images as ordered input channels. For each sample, all pick-up demand images at the selected historical time lags are stacked first, followed by the corresponding drop-off demand images at the same lags. Fig. \ref{fig:input_representation_flowchart} illustrates the pick-up case for clarity; the same construction is extended to drop-off demand, and the network produces a two-channel output representing pick-up and drop-off demand.

Let \( n \) denote the number of historical time lags included per demand type. The resulting input tensor is defined as:
\begin{equation}
\mathbf{X} \in \mathbb{R}^{C \times H \times W}, \quad C = 2n
\end{equation}
where \( H \) and \( W \) denote the height and width of the image in pixels, and \( C \) denotes the number of image channels with a factor of two to account for pick-up and drop-off channels.

To stabilize training, all input demand images are normalized using a global scaling factor equal to the maximum pixel intensity observed in the dataset, while the target pick-up and drop-off images remain in their original scale so that the model predicts absolute demand values. The specific historical time lags are determined by the selection procedure described in Section \ref{optimal-depth}; this subsection focuses solely on the representation and construction of the input tensor once the time-lag set has been defined.

Through this spatiotemporal channel construction, the proposed input representation preserves fine-grained spatial demand patterns within each image while encoding temporal dynamics through the ordered channel sequence, enabling effective learning of short-term demand evolution using convolutional deep learning models.

\subsection{Deep Learning Model}
\label{subsec:deep-learning-model}
\noindent To empirically validate the proposed temporal input structure, a standard UNET encoder-decoder architecture \cite{UNet2015Convolutional} is employed as the demand prediction model in this study. UNET provides a strong image-to-image regression benchmark for grid-structured demand data. This choice is supported by our earlier benchmarking study \cite{Sahnoon2024UNETUNETR}, in which both UNET \cite{UNet2015Convolutional} and transformer-augmented UNETR \cite{Hatamizadeh2022UnetrTransformers} outperformed the published MFCN model \cite{Phithakkitnukooon2021PredictingSpatiotemporal} across most demand levels and under a consistent experimental setup. The MFCN study also showed that the MFCN benchmark outperformed other baselines such as na\"ive forecasting, linear regression \cite{Montgomery2012IntroductionLinear}, and convolutional long short-term memory (LSTM) models \cite{Ham2021SpatiotemporalDemand}. Based on this evidence, the present work adopts a standard UNET as a stable and empirically validated baseline and avoids architectural modifications so that performance differences can be attributed to temporal input design rather than network complexity.

The adopted UNET follows a conventional encoder-decoder structure with skip connections. Each encoder stage applies two $3\,\times\,3$ padded convolutions, each followed by batch normalization and ReLU activation, and uses $2\,\times\,2$ max pooling for downsampling. The network starts with 64 feature channels and doubles the channel width after each downsampling step. The decoder mirrors this hierarchy by upsampling the feature maps using bilinear upsampling, concatenating them with the corresponding encoder features, and applying two additional $3\times3$ convolution layers, each followed by batch normalization \cite{Ioffe2015BatchNormalization} and ReLU activation. A final $1\,\times\,1$ convolution produces a two-channel output representing pick-up and drop-off demand for each grid cell. Because the task is regression rather than segmentation, a ReLU activation is applied after the final $1\,\times\,1$ convolution to enforce non-negative demand predictions. All model parameters are learned end-to-end using masked MSE loss computed over the historically active spatial grid cells only, as defined by the global binary activity mask. This masking ensures that training and evaluation focus exclusively on spatial regions with meaningful demand activity and avoids artificially deflating error values through inclusion of inactive pixels.

Training is performed using the root mean square propagation (RMSprop) optimizer \cite{Hinton2012RmspropDivide} with a learning rate of $1\times10^{-5}$ and batch size $4$, which provides stable convergence for this architecture. Models are trained for a maximum of $50$ epochs with early stopping based on validation loss using a patience of $6$ epochs. A ReduceLROnPlateau scheduler is applied with a reduction factor of $0.5$, patience of $2$ epochs, and a minimum learning rate of $1\times10^{-7}$. The scheduler is triggered based on relative improvements in validation loss to avoid reacting to minor fluctuations. All experiments in this study use the same architecture and training settings so that performance differences reflect temporal input design rather than model capacity or optimization changes.

\subsection{Loss Function}
\label{subsec:loss_function}

\noindent Let \(Y \in \mathbb{R}^{2 \times H \times W}\) denote the ground-truth demand image for a single sample, where the two channels correspond to pick-up and drop-off demand, and \(H \times W\) denote the height and width of the demand image. Let \(\widehat{Y} \in \mathbb{R}^{2 \times H \times W}\) denote the corresponding model prediction. Let \(M \in \{0,1\}^{H \times W}\) be the global binary activity mask over the spatial grid, indexed by row \(r\) and column \(c\), where \(M_{rc}=1\) indicates an active spatial cell (i.e., pixel) within the valid operating region and \(M_{rc}=0\) otherwise. The mask is applied identically to both output channels. The masked mean squared error (masked MSE) for a single sample is computed as the average squared error over all masked spatial cells in both output channels (i.e., \(C=2\)) and is defined as:
\begin{equation}
\mathcal{L}_{\mathrm{mMSE}}(\widehat{Y},Y)
=
\frac{\sum_{k=1}^{2}\sum_{r=1}^{H}\sum_{c=1}^{W}
M_{rc}\left(\widehat{Y}_{krc}-Y_{krc}\right)^2}
{2\sum_{r=1}^{H}\sum_{c=1}^{W} M_{rc}}
\end{equation}

For a training batch $\mathcal{B}$, the loss is computed as the average of the per-sample masked losses:
\begin{equation}
\mathcal{L}_{\mathcal{B}}
=
\frac{1}{|\mathcal{B}|}
\sum_{n \in \mathcal{B}}
\mathcal{L}_{\mathrm{mMSE}}
\!\left(\widehat{Y}^{(n)},Y^{(n)}\right)
\end{equation}

This formulation ensures that optimization focuses exclusively on spatial locations with meaningful demand activity while preventing inactive grid cells (i.e., pixels) from artificially reducing the training objective.

\subsection{Evaluation Metrics}
\label{subsec:Evaluation-metrics}

\noindent All trained models are evaluated on the held-out test subset described in Section \ref{subsec:dataset_splitting}. For each test sample, models produce two output channels corresponding to pick-up and drop-off demand, and all reported metrics are averaged jointly over both channels using the fixed global binary activity mask defined in Section \ref{subsec:global-mask} to reflect predictive performance exclusively within spatially meaningful demand regions.

\subsubsection{Masked error-based metrics}

Following the notation introduced in Section \ref{subsec:loss_function}, masked MSE is reused during evaluation as the primary error metric. In addition, complementary masked error metrics are computed to capture different aspects of predictive performance. The masked MAE is defined as:
\begin{equation}
\mathrm{MAE} =
\frac{
\sum_{k=1}^{2}\sum_{r=1}^{H}\sum_{c=1}^{W}
M_{rc}
\left| \widehat{Y}_{krc} - Y_{krc} \right|
}{
2\sum_{r=1}^{H}\sum_{c=1}^{W} M_{rc}
}
\end{equation}

The masked maximum absolute error (MaxAE), reflecting the largest deviation within the active region, is defined as:
\begin{equation}
\mathrm{MaxAE} =
\max_{\substack{k \in \{1,2\} \\ r,c \,:\, M_{rc}=1}}
\left| \widehat{Y}_{krc} - Y_{krc} \right|
\end{equation}

\noindent These metrics provide complementary perspectives, where MAE reflects average predictive accuracy and MaxAE highlights extreme local errors that are particularly relevant for operational robustness.

\subsubsection{Masked coefficient of determination}

A masked coefficient of determination ($R^2$) is computed on a per-sample basis and averaged across the test subset. Using the same notation, the masked mean per-sample $R^2$ is given by:
\begin{equation}
R^2 =
1 -
\frac{
\sum_{k=1}^{2}\sum_{r=1}^{H}\sum_{c=1}^{W}
M_{rc}
\left( Y_{krc} - \widehat{Y}_{krc} \right)^2
}{
\sum_{k=1}^{2}\sum_{r=1}^{H}\sum_{c=1}^{W}
M_{rc}
\left( Y_{krc} - \bar{Y}_{k} \right)^2
+ \varepsilon
}
\end{equation}
where $\bar{Y}_{k}$ denotes the mean of $Y_{krc}$ over masked pixels for channel $k$, and $\varepsilon$ is a small constant introduced to stabilize the denominator when the variance within the masked region becomes extremely small. This metric imposes a strict criterion on sparse-demand frames, penalizing predictions that introduce spatial structure when the ground-truth exhibits little or no spatial variability within the active region. Across all evaluations, mean per-sample $R^2$ is therefore interpreted comparatively and in conjunction with error-based metrics, as its sensitivity to within-mask variance varies across demand regimes and prediction horizons.

\subsection{Statistical Testing}
\label{subsec:statistical_testing}

\noindent To determine whether observed performance differences between temporal input configurations are statistically meaningful, paired statistical testing is conducted using per-sample masked MSE distributions. The subset used for statistical testing depends on the evaluation stage.

During temporal depth selection, statistical comparisons are performed on the per-sample masked MSE results computed on the validation subset. In this stage, multiple candidate temporal depth configurations are trained using the same training data and compared through their validation-set performance distributions. This ensures that optimal temporal depth selection is based solely on validation performance, preventing data leakage and preserving the test subset for unbiased final evaluation.

For the final comparison between the proposed temporal input structure and alternative historical lag set methodologies, statistical testing is conducted on the per-sample masked MSE results computed on the test subset. In this final reporting stage, the selected configuration is compared against competing configurations through their test-set performance distributions, providing an unbiased assessment of generalization performance.

Since all configurations are evaluated against identical ground-truth targets within a given subset, statistical testing is performed on the paired per-sample MSE differences. The normality of the paired differences is assessed using the Shapiro-Wilk test \cite{SHAPIRO1965VarianceTest}. Given the relatively large sample size, Shapiro-Wilk becomes highly sensitive to even minor deviations from normality. In practice, paired MSE differences also exhibit skewness and mild heteroskedasticity. Consequently, the normality assumption is rejected in all reported comparisons, and non-parametric Wilcoxon signed-rank tests \cite{Wilcoxon1945IndividualComparisons} are adopted.

Two complementary testing strategies are employed. First, superiority testing is conducted using two-sided Wilcoxon signed-rank tests to determine whether performance differences are statistically significant. Second, when assessing whether a simpler configuration can be considered practically comparable to a reference model, a non-inferiority (NI) analysis is performed using a one-sided Wilcoxon signed-rank test applied to shifted paired differences under a predefined margin.

To control the family-wise error rate associated with multiple comparisons, Holm's \cite{Holm1979SimpleSequentially} step-down correction is applied with a significance level of $\alpha = 0.05$. A result is considered statistically significant when the Holm-corrected $p$-value is less than $\alpha$. This two-stage testing framework ensures that temporal depth selection is performed without exposure to test data, while final reported improvements reflect statistically reliable generalization performance on unseen samples.

\section{Temporal Input Design for Deep Learning}
\label{sec:temporal_input_design}
\noindent This section builds upon the experimental setup defined in Section \ref{sec:exp-design} and presents the proposed methodology for designing the temporal input structure provided to the demand prediction model. First, informative historical time lags are identified using a combined Pearson correlation- and error-based analysis. Next, the methodology used to determine the optimal set of historical time lags is described through a controlled ablation study based on a fixed UNET model configuration, supported by paired non-parametric statistical testing and Holm correction. Finally, baseline temporal input configurations are introduced for comparative evaluation against our proposed temporal input design.

\subsection{Selection of Informative Historical Time Lags Using Pearson Correlation and Error Analysis}
\label{subsec:lag_selection}

\noindent The selection of historical demand time steps is guided by a combined correlation- and error-based analysis designed to identify which past demand patterns are the most informative to predict future demand. This analysis is conducted on the training subset independently for next-hour and next 24-hour demand prediction tasks, while respecting the maximum allowable lookback window of 504 hours established during dataset splitting.

For each target time \(t\), the corresponding pick-up and drop-off demand images are denoted as \(I_{p}(t)\) and \(I_{d}(t)\), respectively. For a candidate historical time lag \(\tau\), the preceding demand images are given by \(I_{p}(t-\tau)\) and \(I_{d}(t-\tau)\). All computations are performed over the valid service area by restricting computation to the active spatial support \(\Omega\).

To quantify the similarity between target demand patterns and their historical counterparts independent of magnitude, we first compute the average same-channel Pearson correlation coefficient:
\begin{equation}
\begin{aligned}[b]
\label{eq:self-corr}
C_s(\tau)
=
\frac{1}{2}
[corr\big(I_{p}(t), I_{p}(t-\tau)\big) + \\corr\big(I_{d}(t), I_{d}(t-\tau)\big)]
\end{aligned}
\end{equation}
where \(corr(x,y)\) denotes the Pearson correlation coefficient between flattened demand images computed over \(\Omega\).
This metric captures similarity in spatial demand structure while remaining invariant to overall demand scale. For completeness, we also compute a cross-channel Pearson correlation metric:
\begin{equation}
\begin{aligned}[b]
\label{eq:cross-corr}
C_c(\tau)
=
\frac{1}{2}
[corr\big(I_{p}(t), I_{d}(t-\tau)\big)+ \\corr\big(I_{d}(t), I_{p}(t-\tau)\big)]
\end{aligned}
\end{equation}
which provides a scale-free measure of pick-up and drop-off coupling. 

To account for magnitude differences between demand images, we next compute the average same-channel MAE over the active spatial support \(\Omega\):
\begin{equation}
\label{eq:self-MAE}
\begin{aligned}
\mathrm{MAE}_s(\tau) = \frac{1}{2|\Omega|} [ \lVert I_{p}(t) - I_{p}(t-\tau) \rVert_1 + \\ \lVert I_{d}(t) - I_{d}(t-\tau) \rVert_1]
\end{aligned}
\end{equation}
where \(\lVert \cdot \rVert_1\) denotes the \(L_1\) norm computed over active grid cells (i.e., image pixels) \((r,c)\in\Omega\). This metric measures the average per-pixel deviation between historical and target demand images and serves as the primary magnitude-based criterion for lag ranking. Cross-channel magnitude-based error is intentionally excluded from the ranking process, as it does not align with the objective of selecting lags that preserve same-channel demand memory.

For diagnostic purposes, we also compute the corresponding same-channel absolute difference sum (AD):
\begin{equation}
\label{eq:self-AD}
\mathrm{AD}_s(\tau)
=
\lVert I_{p}(t) - I_{p}(t-\tau) \rVert_1
+
\lVert I_{d}(t) - I_{d}(t-\tau) \rVert_1
\end{equation}
Because the historically active spatial support \(\Omega\) and image resolution are fixed, \(\mathrm{AD}_s(\tau)\) is proportional to \(\mathrm{MAE}_s(\tau)\) up to a constant factor and is therefore used only as a diagnostic measure rather than in the ranking procedure. More generally, in frameworks where the spatial support varies over time (e.g., due to dynamic masking or time-varying service regions), \(\mathrm{AD}_s(\tau)\) may provide complementary information and can be included in the ranking procedure, subject to a redundancy check with \(\mathrm{MAE}_s(\tau)\).
In such cases, when strong collinearity is detected, \(\mathrm{MAE}_s(\tau)\) is preferred due to its scale-invariant interpretation.

For each candidate lag \(\tau\), the metrics defined above are averaged across all valid target instances to obtain summary statistics. Candidate lags are then ordered using a weight-free rank aggregation procedure to avoid introducing arbitrary scaling or weighting across heterogeneous metrics.

For each retained metric \(m\), a per-metric rank \(\mathrm{rank}_m(\tau)\) is assigned to every lag \(\tau\). For correlation-based metrics (\(C_s\) and \(C_c\)), lags are ranked in descending order such that higher correlation values receive lower (better) ranks. For the magnitude-based metric (\(\mathrm{MAE}_s\)), lags are ranked in ascending order such that lower error values receive lower (better) ranks.
Average ranking is used to handle ties. Formally, the per-metric rank is defined as:
\begin{equation}
\label{eq:rankm}
\mathrm{rank}_m(\tau)=
\begin{cases}
\mathrm{rank}\big(-S_m(\tau)\big), & \text{if } m \in \{ C_s,\, C_c \} \\
\mathrm{rank}\big(S_m(\tau)\big),  & \text{if } m = \mathrm{MAE}_s
\end{cases}
\end{equation}

where \(S_m(\tau)\) denotes the aggregated score of metric \(m\) at lag \(\tau\), and \(\mathrm{rank}(\cdot)\) assigns rank 1 to the best-performing value.

An aggregate ranking score for each lag is then computed as:
\begin{equation}
\label{eq:rank-avg}
\mathrm{Rank}_{\mathrm{Avg}}(\tau)
=
\frac{1}{K}
\sum_{m=1}^{K}
\mathrm{rank}_m(\tau)
\end{equation}
where \(K\) denotes the number of metrics included in the aggregation. Lower values of \(\mathrm{Rank}_{\mathrm{Avg}}(\tau)\) indicate more informative historical time lags. The final discrete ranking \(\mathrm{Rank}_{\mathrm{final}}(\tau)\) is obtained by sorting \(\mathrm{Rank}_{\mathrm{Avg}}(\tau)\) in ascending order and assigning integer ranks; ties are resolved by favoring lags with lower \(\mathrm{MAE}_s\) variance, indicating more stable predictive performance.

Applying this procedure within the 504-hour window yields the top 20 historical time lags for each prediction horizon. These lags, expressed relative to the target time \(t\) as \((t-\tau)\), are summarized in Table \ref{table:selected_lags} along with their final ranks $\mathrm{Rank_{final}}(\tau)$, and apply to both pick-up and drop-off demand.

The selected lags exhibit a clear and interpretable temporal structure. For the next-hour prediction task, selected lags include immediate past demand at \((t-1)\), \((t-2)\), and \((t-3)\), capturing short-term temporal persistence. These are followed by clusters of lags centered around the daily cycle, including \((t-23)\), \((t-24)\), and \((t-25)\), which correspond to the same hour on the previous day and its immediate neighboring hours. Additional groups of lags appear around longer periodic intervals, particularly at \((t-168)\), \((t-336)\), and \((t-504)\), with several adjacent hours selected around each reference point. These clusters reflect recurring weekly demand structures rather than isolated temporal offsets.

A similar pattern is observed for the next 24-hour prediction task, where dominant lags cluster around multiples of 24 hours, such as \((t-24)\) and \((t-48)\), followed by repeated weekly recurrence at \((t-168)\), \((t-336)\), and \((t-504)\). Together, these patterns confirm that both short-term temporal continuity and cyclical daily and weekly behaviors play a central role in predicting e-scooter demand, while highlighting the importance of balancing temporal coverage with model complexity.

\begin{table}[t]
\centering
\caption{Top 20 selected historical time lags (in hours) for next-hour and next 24-hour e-scooter demand prediction. Each lag $\tau$ corresponds to demand observed at time $(t-\tau)$.}
\label{table:selected_lags}
\begin{tabular}{c | c||c | c}
\hline
\multicolumn{2}{c||}{\textbf{Next-hour}} & \multicolumn{2}{c}{\textbf{Next 24-hour}} \\
\hline
\textbf{Rank\textsubscript{final}($\tau$)} & \textbf{$\tau$ (hours)} & \textbf{Rank\textsubscript{final}($\tau$)} & \textbf{$\tau$ (hours)} \\
\hline
\hline
1  & 1   & 1  & 24  \\
4  & 2   & 3  & 25  \\
17 & 3   & 10 & 26  \\
11 & 22  & 9  & 48  \\
5  & 23  & 12 & 143 \\
2  & 24  & 6  & 144 \\
6  & 25  & 11 & 145 \\
14 & 26  & 18 & 166 \\
13 & 48  & 5  & 167 \\
16 & 143 & 2  & 168 \\
9  & 144 & 4  & 169 \\
15 & 145 & 20 & 170 \\
8  & 167 & 17 & 191 \\
3  & 168 & 8  & 192 \\
7  & 169 & 16 & 193 \\
12 & 192 & 19 & 312 \\
20 & 335 & 15 & 335 \\
10 & 336 & 7  & 336 \\
19 & 337 & 14 & 337 \\
18 & 504 & 13 & 504 \\
\hline
\end{tabular}
\end{table}

\subsection{Determining the Optimal Set of Historical Time Lags for Model Input}
\label{optimal-depth}
\noindent After ranking the informative historical time lags, the next step is to determine how many of them should be provided to the deep learning model. Beyond identifying the configuration that minimizes validation-set MSE, we also assess whether a simpler configuration can be considered non-inferior under a predefined margin, allowing computational savings without meaningful loss of predictive accuracy.

To this end, a controlled ablation study is conducted using the UNET architecture. For each prediction horizon, models are trained on the same training subset while varying the total number of input demand maps, where each map corresponds to a pick-up or drop-off demand image at a selected historical time lag. Thus, a 40-channel model contains 20 pick-up and 20 drop-off historical demand images.

For each horizon, training begins with the top 20 ranked time lags for each demand type, yielding 40 input channels. The number of channels is then reduced in steps of two by removing, at each step, the lowest-ranked lag for both pick-up and drop-off according to Table \ref{table:selected_lags}. This produces candidate models ranging from 40 to 4 input channels while keeping all architecture and training settings fixed. All candidate models are then evaluated on the validation subset using per-sample masked MSE, while the test subset is reserved exclusively for the final comparison to prevent data leakage.

To enable rigorous model comparison, paired statistical tests are performed on the per-sample MSE distributions. Since all temporal input configurations are evaluated against identical ground-truth targets, statistical testing is conducted on the paired per-sample MSE differences. The normality of these paired differences is assessed using the Shapiro-Wilk test. In all reported comparisons, the normality assumption is violated; therefore, non-parametric Wilcoxon signed-rank tests are adopted.

Two types of statistical evaluation are conducted. First, superiority testing is performed using two-sided Wilcoxon signed-rank tests. This includes both exhaustive pairwise comparisons across all candidate configurations and focused comparisons against the minimum-MSE model to identify statistically superior alternatives. 

Second, a non-inferiority (NI) assessment is performed to determine whether a simpler temporal configuration can be considered practically indistinguishable from the minimum-MSE model within a predefined tolerance. Let $\mathrm{MSE}_{i}^{(\mathrm{cand})}$ and $\mathrm{MSE}_{i}^{(\min)}$ denote the per-sample MSE values for evaluation sample $i$ obtained from the candidate configuration and the minimum-MSE configuration, respectively, where $i=1,\ldots,N$ indexes the evaluation samples. The NI analysis evaluates the shifted paired differences:

\begin{equation}
d_i = \mathrm{MSE}_{i}^{(\mathrm{cand})} - \mathrm{MSE}_{i}^{(\min)} - \delta
\end{equation}

\noindent where the non-inferiority margin $\delta$ is defined as:

\begin{equation}
\delta = 0.02\,\overline{\mathrm{MSE}}_{\min}
\end{equation}

\noindent where $\overline{\mathrm{MSE}}_{\min}$ denotes the mean per-sample MSE of the minimum-MSE configuration. 

A one-sided paired Wilcoxon signed-rank test is then applied to assess whether the shifted paired differences are statistically less than or equal to zero, indicating that the candidate configuration does not exceed the predefined non-inferiority margin. To control the family-wise error rate across multiple comparisons, Holm's step-down correction is applied throughout with a significance level of $\alpha = 0.05$.

\subsection{Baseline Temporal Input Configurations for Comparative Evaluation}
\noindent For each prediction horizon, two alternative historical time-lag configurations are considered as baselines for comparison with the proposed time-lag selection method: a recent adjacent history baseline and a fixed-period history baseline. Recent adjacent historical time lags have been used in prior micromobility demand prediction studies that rely on previous hourly demand of the same day or consecutive recent hours to predict future demand \cite{Ham2021SpatiotemporalDemand, Kim2022PredictingDemand, Zhang2022FreeFloating}. Prior work has also shown the relevance of periodic historical demand information, either through simple same-hour previous-day baselines \cite{Zhang2022FreeFloating} or through hybrid temporal input designs that combine recent observations with daily and weekly recurrence \cite{Li2023ImprovingShort, Nejadshamsi2026PredictingShort, Hosseinpanahi2026MachineLearning}. Motivated by these patterns, the two baselines selected here separate the recent-adjacent and periodic components of historical demand, allowing each to be evaluated independently under the same controlled modelling setting. The time-lag set associated with each configuration and prediction horizon is summarized in Table \ref{table:tau_configurations}, where the number of input channels for each horizon is aligned with the optimal temporal depth (i.e., minimal non-inferior) identified later in Section \ref{subsec:val-of-optimal-lags}. In all cases, the same set of time lags is applied identically to both pick-up and drop-off demand, resulting in separate input channels with a shared temporal structure. These baseline configurations can be described as follows:

\begin{enumerate}
\item \textbf{Recent Adjacent history baseline:}
This configuration assumes that the most recent observations immediately preceding the target prediction time contain the most relevant information for demand prediction. Consequently, the model input consists of an adjacent block of recent historical observations, defined separately for each prediction horizon. For next-hour prediction, the input captures short-term temporal persistence by using the most recent consecutive hours prior to the target time. For next 24-hour prediction, the configuration is shifted to align with the same hour of the previous day and its immediately preceding consecutive hours, reflecting short-term temporal continuity at a daily scale.

\item \textbf{Fixed-period historical baseline:}
This configuration assumes that demand exhibits strong periodic structure tied to the hour of the day. The model input therefore consists of observations taken at fixed periodic intervals aligned with a consistent temporal offset relative to the target time, and is applied identically to pick-up and drop-off demand. For next-hour prediction, this configuration includes demand observed at the previous hour relative to the target time \(t\) (i.e., \(t-1\)), together with the same relative hour across multiple preceding days (i.e., \((t-1)-24d\) for several values of \(d\)). For next 24-hour prediction, the configuration captures pure daily periodicity by selecting the same hour of the previous day across multiple prior days.

\end{enumerate}

\begin{table*}[!t]
\centering
\caption{Historical Time-lag configurations used to validate the proposed historical time-lag selection method under the next-hour and next 24-hour prediction horizons. For each configuration, the same set of time lags \( \tau \) is applied identically to pick-up and drop-off demand, resulting in separate input channels with a shared temporal structure.}
\label{table:tau_configurations}
\begin{tabular}{c l c l}
\hline
\textbf{Prediction Horizon} & \textbf{Configuration Type} & \textbf{Channels} & \textbf{Time-lag set \( \tau \) (hours)} \\
\hline
\hline
\multirow{4}{*}{Next-hour}
& Proposed method
& 36
& \(\{1, 2, 3, 22, 23, 24, 25, 26, 48, 143, 144, 145, 167, 168, 169, 192, 336, 504\}\) \\

& Recent adjacent history
& 36
& \(\{1, 2, 3, \ldots, 17, 18\}\) \\

& \multirow{2}{*}{Fixed-period history}
& \multirow{2}{*}{36}
& \(\{1, 25, 49, 73, 97, 121, 145, 169, 193, 217, 241, 265, 289, 313, 337, 361,\) \\

&  &  &
\(385, 409\}\) \\
\hline
\multirow{3}{*}{Next 24-hour}
& Proposed method
& 18
& \(\{24, 25, 48, 144, 167, 168, 169, 192, 336\}\) \\

& Recent adjacent history
& 18
& \(\{24, 25, 26, 27, 28, 29, 30, 31, 32\}\) \\

& Fixed-period history
& 18
& \(\{24, 48, 72, 96, 120, 144, 168, 192, 216\}\) \\
\hline
\end{tabular}
\end{table*}

\section{Results}
\label{sec:results}
\noindent This section presents the empirical results of the proposed historical time-lag selection framework in three stages. First, using the validation subset, we identify for each prediction horizon the historical time-lag configuration that achieves the minimum MSE and determine the corresponding minimal non-inferior configuration through Holm-corrected non-inferiority testing. Second, these selected temporal input structures are evaluated against alternative baseline time-lag configurations on the held-out test subset using a controlled experimental design in which the same UNET demand prediction model is trained for each temporal input configuration while all other modelling components are held fixed. Third, the proposed temporal input design is statistically validated through pairwise significance testing to assess whether the observed performance differences relative to the baseline configurations are statistically meaningful.

\subsection{Validation-Based Selection of Optimal Historical Time Lags}
\label{subsec:val-of-optimal-lags}
\noindent This subsection reports the validation-subset results used to determine the optimal set of historical time lags for each prediction horizon. Specifically, we identify the minimum-MSE configuration and assess whether a simpler configuration is minimal non-inferior under Holm-corrected non-inferiority testing. Table \ref{table:optimal_temporal_lags_set} summarizes the resulting minimum-MSE and minimal non-inferior configurations selected using the UNET model. This analysis establishes the final temporal input depth used for subsequent comparative evaluation.

\subsubsection{Next-hour prediction}
For the next-hour prediction horizon, the minimum MSE is achieved by the 36-channel model, corresponding to 18 pick-up and 18 drop-off historical demand images. This model serves as the reference winner for subsequent comparisons. Holm-corrected paired testing indicates that no model with fewer input channels satisfies the predefined non-inferiority criterion relative to the 36-channel model. Consequently, the 36-channel configuration is selected as both the optimal and minimal non-inferior temporal depth for next-hour prediction. This result suggests that short-term demand prediction benefits from incorporating a relatively rich temporal context, and that reducing the temporal depth below this level leads to statistically significant degradation in performance.

\subsubsection{Next 24-hour prediction}
For the next 24-hour prediction horizon, the minimum MSE is attained by the 26-channel model, corresponding to 13 pick-up and 13 drop-off historical demand images. However, Holm-corrected non-inferiority testing indicates that a simpler 18-channel model, using 9 pick-up and 9 drop-off time lags, satisfies the predefined 2\% non-inferiority margin relative to the minimum-MSE model. As a result, the 18-channel model is selected as the minimal non-inferior temporal depth for next 24-hour prediction.

Overall, these results indicate that the optimal temporal depth depends on the prediction horizon. While next-hour prediction requires a denser set of historical input to capture short-term dynamics, next 24-hour prediction exhibits diminishing returns beyond a certain number of time lags. This horizon-dependent behavior motivates the use of distinct temporal configurations in subsequent experiments, balancing predictive performance with computational efficiency.

\begin{table*}[!t]
\centering
\caption{Optimal historical input configurations identified for each prediction horizon using the UNET model. The table reports the validation-set minimum-MSE configuration and the minimal non-inferior configuration determined through Holm-corrected non-inferiority testing.}
\label{table:optimal_temporal_lags_set}
\begin{tabular}{ccccc}
\hline
\textbf{Prediction Horizon} & \textbf{Configuration} & \makecell{\textbf{Input}\\\textbf{Channels}} & \textbf{MSE*} & \makecell{\textbf{Time Lags ($\tau$)}\\\textbf{(See Table \ref{table:selected_lags})}} \\
\hline
\hline

\multirow{2}{*}{Next-hour} 
& Minimum MSE 
& 36 
& 20.11 
& Top 18 \\

& Minimal non-inferior 
& \multicolumn{3}{c}{Same as minimum-MSE configuration} \\

\hline

\multirow{2}{*}{Next 24-hour} 
& Minimum MSE 
& 26 
& 61.20 
& Top 13 \\

& Minimal non-inferior 
& 18 
& 63.42 
& Top 9 \\

\hline
\end{tabular}

\vspace{2mm}
\footnotesize{*MSE values represent the mean per-sample MSE over the validation subset. 
Non-inferiority testing is performed on paired per-sample MSE differences with a non-inferiority margin of $2\%$ relative to the minimum-MSE configuration.}
\end{table*}

\subsection{Comparative Evaluation Against Baseline Temporal Input Configurations}
\noindent Table \ref{table:val_metrics_next_hour} and Table \ref{table:val_metrics_next_24h} summarize the overall test-set performance of each temporal input configuration using mean per-sample masked MSE, MAE, \(R^2\), and MaxAE, averaged jointly over pick-up and drop-off demand within the global binary activity mask. For next-hour prediction, the model trained using the proposed time-lag selection method achieves the lowest MSE (8.45) and MAE (0.50), the highest mean per-sample $R^2$ score (0.78), and the lowest MaxAE (24.33), clearly outperforming both the recent adjacent history and the fixed-period history baselines across all reported metrics. The recent adjacent history configuration ranks second (MSE = 11.40, $R^2$ = 0.57), while the fixed-period history baseline exhibits the highest error and lowest $R^2$ among the three configurations (MSE = 13.48, $R^2$ = 0.55), indicating a consistent performance ordering aligned with the proposed method.

For next 24-hour prediction, the proposed method again yields the lowest error values (MSE = 26.37, MAE = 0.75, MaxAE = 50.47), demonstrating substantial improvement relative to both alternative historical lag configurations. Although the mean per-sample $R^2$ values are negative for all configurations at this longer prediction horizon, the fixed-period history baseline attains the highest (least negative) score ($R^2$ = $-0.36$), followed by the proposed method ($R^2$ = $-0.77$) and the recent adjacent history baseline ($R^2$ = $-1.71$). In this context, negative mean per-sample $R^2$ values indicate that, on average, the squared prediction error within the masked active region exceeds that of a per-sample baseline predictor that outputs the masked spatial mean of the ground-truth demand. This behavior reflects the increased difficulty of longer-horizon demand prediction under predominantly sparse-demand conditions. Nevertheless, the consistent reduction in MSE, MAE, and MaxAE confirms that the proposed temporal input configuration provides the strongest overall predictive accuracy across both prediction horizons.

\begin{table}[!t]
\centering
\caption{Overall test-set performance of temporal input (time-lag) selection methodologies for next-hour e-scooter demand prediction using the UNET model, computed within the masked active region and reported as metrics averaged jointly over pick-up and drop-off demand channels.}
\label{table:val_metrics_next_hour}
\begin{tabular}{l c c c c}
\hline
\textbf{Configuration} & \textbf{MSE} & \textbf{MAE} & \textbf{$R^2$} & \textbf{MaxAE} \\
\hline
\hline
Proposed method & \textbf{8.45} & \textbf{0.50} & \textbf{0.78} & \textbf{24.33} \\
Recent adjacent history & 11.40 & 0.57 & 0.57 & 34.51 \\
Fixed-period history & 13.48 & 0.61 & 0.55 & 38.03 \\
\hline
\end{tabular}
\end{table}

\begin{table}[!t]
\centering
\caption{Overall test-set performance of temporal input (time-lag) selection methodologies for next 24-hour e-scooter demand prediction using the UNET model, computed within the masked active region and reported as metrics averaged jointly over pick-up and drop-off demand channels.}
\label{table:val_metrics_next_24h}
\begin{tabular}{l c c c c}
\hline
\textbf{Configuration} & \textbf{MSE} & \textbf{MAE} & \textbf{$R^2$} & \textbf{MaxAE} \\
\hline
\hline
Proposed method & \textbf{26.37} & \textbf{0.75} & -0.77 & \textbf{50.47} \\
Recent adjacent history & 40.46 & 0.87 & -1.71 & 64.46 \\
Fixed-period history & 30.86 & 0.85 & \textbf{-0.36} & 53.26 \\
\hline
\end{tabular}
\end{table}

\subsection{Statistical Validation of the Proposed Temporal Input Structure}
\noindent Statistical significance is assessed using the paired testing framework described in Section \ref{subsec:statistical_testing}, applied to per-sample masked MSE distributions on the test subset. Table \ref{table:config_test_holm} summarizes the Holm-corrected pairwise comparison results for both prediction horizons. For next-hour prediction, the model trained using the proposed time-lag selection method achieves significantly lower MSE than both the recent adjacent history baseline and the fixed-period history baseline. In addition, the recent adjacent history baseline also significantly outperforms the fixed-period history baseline, indicating a clear performance ordering among the three configurations.

For next 24-hour prediction, the proposed time-lag selection method again achieves significantly lower MSE than both alternative lag configurations, confirming its superiority at the longer prediction horizon. However, the relative ordering of the two baseline configurations differs from the next-hour case where the fixed-period history baseline now significantly outperforms the recent adjacent history baseline. This reversal indicates that pure daily periodic information becomes more informative than short-term adjacent history when predicting the demand at a 24-hour horizon. In all comparisons, the Holm-corrected $p$-values are well below the significance threshold, confirming that the observed performance differences are statistically robust and not attributable to random variation.

\begin{table*}[!t]
\centering
\caption{Holm-corrected pairwise statistical comparison between historical time-lag selection methodologies on test MSE for next-hour and next 24-hour e-scooter demand prediction. Results are based on two-sided Wilcoxon signed-rank tests applied to per-sample MSE distributions averaged jointly over pick-up and drop-off demand.}
\label{table:config_test_holm}
\begin{tabular}{c l l c}
\hline
\textbf{Horizon} & \textbf{Comparison} & \textbf{Better Configuration} & \textbf{Holm corrected\ p-value} \\
\hline
\hline
\multirow{3}{*}{Next-hour}
& Proposed method vs.\ recent adjacent history
& Proposed method
& \(5.78\times10^{-40}\) \\

& Proposed method vs.\ fixed-period history
& Proposed method
& \(1.43\times10^{-75}\) \\

& Recent adjacent vs.\ fixed-period history
& Recent adjacent history
& \(1.89\times10^{-4}\) \\
\hline
\multirow{3}{*}{Next 24-hour}
& Proposed method vs.\ recent adjacent history
& Proposed method
& \(1.15\times10^{-35}\) \\

& Proposed method vs.\ fixed-period history
& Proposed method
& \(1.55\times10^{-8}\) \\

& Recent adjacent vs.\ fixed-period history
& Fixed-period history
& \(2.53\times10^{-32}\) \\
\hline
\end{tabular}
\end{table*}

\section{Discussion}
\label{sec:discussion}
\noindent The results indicate that temporal input design is a critical determinant of predictive performance in deep-learning-based e-scooter demand prediction, even when the underlying network architecture is held fixed. The selected lag configurations consistently combine (i) very recent history (e.g., 1-3 hours for next-hour prediction) with (ii) strong periodic recurrences aligned with daily and weekly travel cycles (e.g., approximately 24, 168, 336, and 504 hours). This structure suggests that short-term demand exhibits immediate persistence, whereas longer-range dependencies are primarily governed by recurring behavioral patterns at daily and weekly intervals. In particular, the optimal next-hour configuration integrates both adjacent-hour context and multiple periodic anchors, indicating that accurate short-term e-scooter demand prediction benefits from simultaneously modelling immediate temporal momentum and recurrent routine dynamics.

The ablation-based temporal depth analysis further demonstrates that the optimal number of historical time lags is horizon-dependent. For next-hour prediction, reducing the temporal depth below 36 channels results in statistically significant performance degradation, implying that short-term dynamics require a relatively rich contextual window to maintain stability across heterogeneous urban regions. In contrast, for next 24-hour prediction, a smaller effective depth is sufficient, reflecting diminishing returns beyond a moderate number of informative periodic and near-periodic lags. From a practical perspective, this finding suggests that longer-horizon e-scooter demand prediction may benefit more from strategically selected periodic structure than from expanding adjacent-hour history.

In the final test-set comparisons, the proposed temporal input design achieves substantial reductions in MSE and MAE across both prediction horizons relative to common heuristic baselines, with these improvements confirmed by Holm-corrected paired testing. For the next 24-hour prediction, the negative $R^2$ values observed across configurations indicate that the task remains challenging relative to a variance-based baseline, likely reflecting increased uncertainty and volatility at extended horizons. Nevertheless, the proposed configuration consistently attains the lowest error-based metrics (MSE and MAE) and exhibits improved robustness as reflected by lower MaxAE values, which is particularly relevant for operational planning during high-demand periods.

\section{Conclusion and Future Work}
\label{sec:conclusion}
\noindent In this paper, we addressed two key gaps in the literature on deep learning for micromobility demand prediction: the lack of a formalized and reproducible framework for transforming raw trip records into model-ready spatiotemporal demand representations, and the limited use of principled, statistically validated methods for selecting historical time lags. 

To address these gaps, this paper presented a grid-based framework for constructing hourly pick-up and drop-off demand images from large-scale shared e-scooter trip data and a statistically validated methodology for designing the temporal input structure used for deep learning demand prediction models. Using the processed Austin, TX 2019 e-scooter trips, we proposed a combined correlation- and error-based approach to identify informative historical time lags and then selected the appropriate temporal depth through a controlled UNET ablation study with paired non-parametric testing and Holm correction. The resulting temporal structures capture both short-term persistence and strong daily and weekly recurrence, leading to optimal depths of 36 input channels (i.e., 18 historical time lags per demand type) for next-hour demand prediction and 18 input channels (i.e., 9 historical time lags per demand type) for next 24-hour demand prediction. In final test-set evaluation, the proposed temporal input design outperformed recent-adjacent and fixed-period lag baselines, achieving markedly lower MSE, MAE, and MaxAE with statistically meaningful improvements.

The findings of this study establish temporal input design as a foundational component of spatiotemporal deep learning for micromobility demand prediction by showing how much predictive performance can be attributed to carefully structured historical time lags, independent of architectural modifications. However, several limitations should be noted. First, the dataset reports trip start and end locations only through census tract identifiers, leading us to assume centroid-based origin and destination locations, which imposes a coarse spatial resolution and contributes to sparsity in the final processed demand representation. Second, the analysis intentionally isolates historical demand information as the sole predictive input; incorporating exogenous drivers such as weather, special events, holidays, and service-availability variables could further enhance long-horizon performance and reduce predictive uncertainty. Finally, the findings are derived from a single city-year dataset; evaluating the proposed methodology across additional cities and time periods would provide stronger evidence of generalization. Despite these limitations, the framework establishes a reproducible and statistically grounded approach to temporal input design that can be extended to other grid-based urban micromobility demand prediction applications.

Future work aimed at further improving predictive performance should therefore build upon this optimized input representation and focus on model architecture design. In particular, architectures that adapt explicitly to distinct demand contexts, such as weekdays, weekends, or other operating regimes, offer promising directions. Examples include day-type conditioning, feature modulation mechanisms, domain generalization techniques, and attention-based spatial and temporal context integration. Such approaches operate after the temporal input stage and may help models better capture heterogeneous demand patterns. Combining such architectural adaptation with the principled temporal input design developed here provides a natural and systematic pathway for advancing micromobility demand prediction.

\bibliography{export}  
\bibliographystyle{IEEEtran}

\newpage
\section{Biography Section}
\begin{IEEEbiography}[{\includegraphics[width=1in,height=1.25in,clip,keepaspectratio]{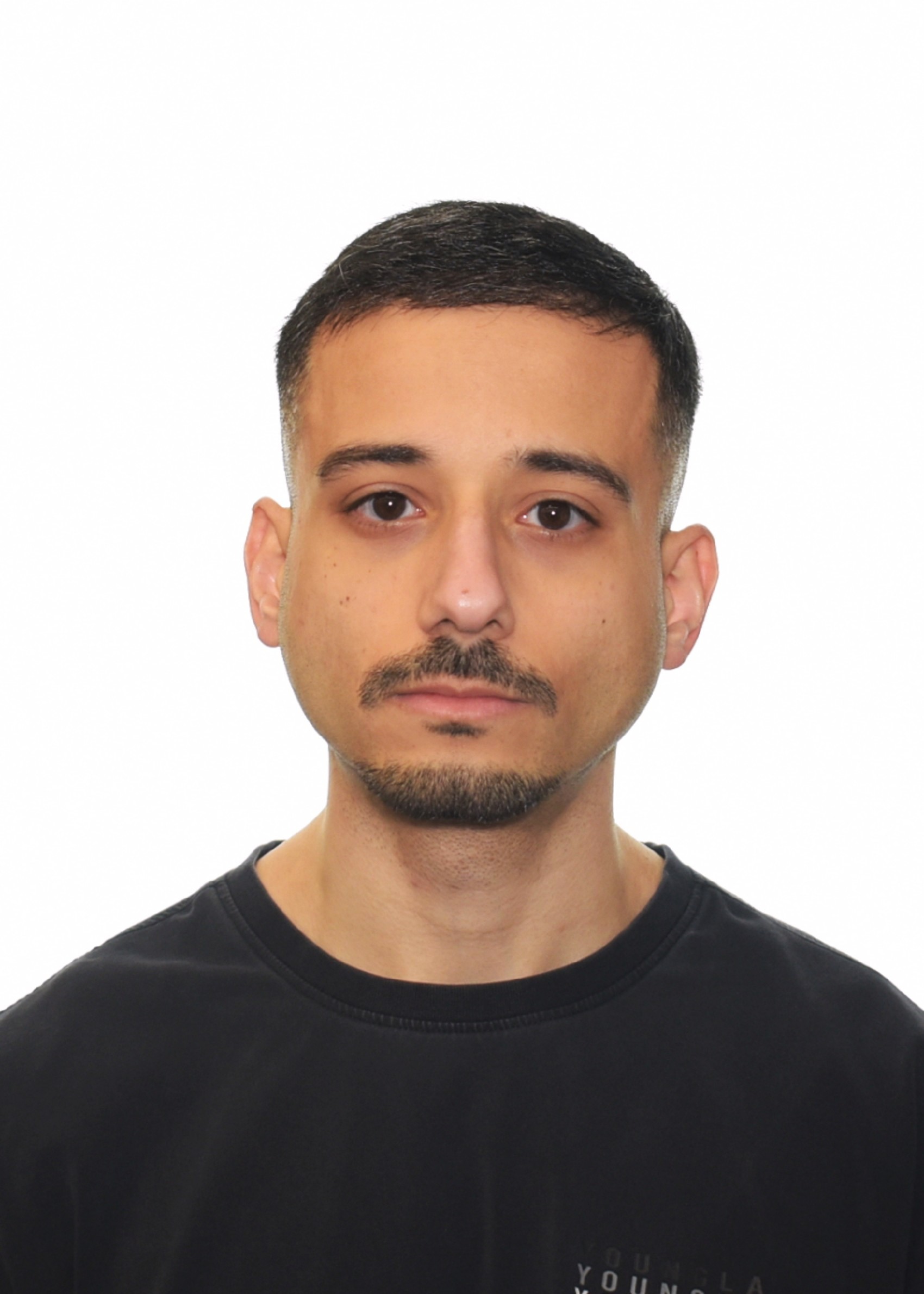}}]{Mohammad Sahnoon}
(Member, IEEE) received the B.Sc. degree in electrical and electronics engineering from the University of Sharjah, Sharjah, UAE, in 2015, and the M.Sc. degree in electrical engineering from the American University of Sharjah, Sharjah, UAE, in 2017.

From 2019 to 2021, he was a Workshop Engineer with the Institute of Applied Technology, Abu Dhabi, UAE. He is currently pursuing the Ph.D. degree in electrical and software engineering at the University of Calgary, Calgary, AB, Canada, where he is also a Graduate Teaching Assistant. His research focuses on intelligent transportation systems, deep learning, spatiotemporal forecasting, and shared micromobility systems, with particular emphasis on demand prediction and data-driven urban mobility modeling.

Mr. Sahnoon is an Engineer-in-Training (E.I.T.) with the Association of Professional Engineers and Geoscientists of Alberta (APEGA).
\end{IEEEbiography}

\vspace{11pt}

\begin{IEEEbiography}[{\includegraphics[width=1in,height=1.25in,clip,keepaspectratio]{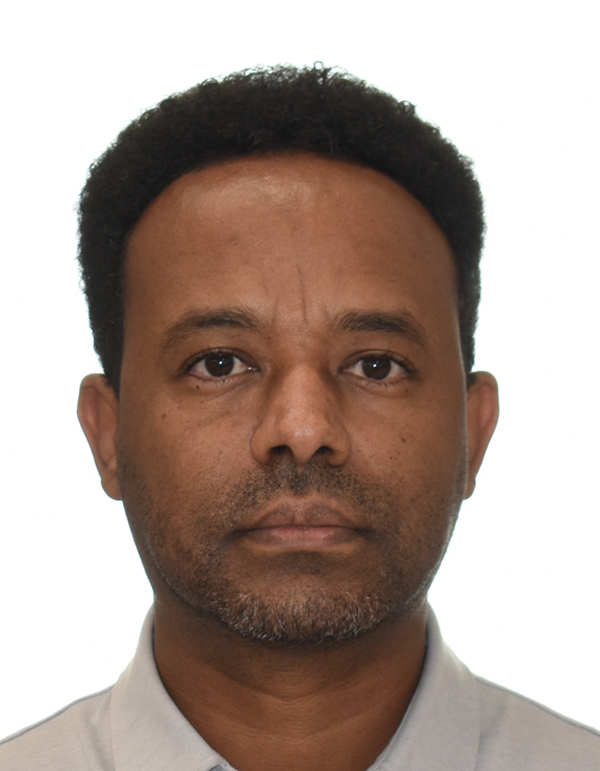}}]{Merkebe Getachew Demissie}
received the M.Sc. degree in transport systems from the Royal Institute of Technology (KTH), Stockholm, Sweden, in 2009, and the Ph.D. degree in transportation systems from the University of Coimbra, Coimbra, Portugal, from the MIT-Portugal program, in 2014. His major field of study is transportation engineering.

He is currently an Assistant Professor in the Department of Civil Engineering at the University of Calgary, Calgary, AB, Canada. He was also a Research Associate at the University of Calgary. Prior to joining the University of Calgary, he was a Postdoctoral Researcher with the Faculty of Sciences and Technology at the University of Coimbra, Coimbra, Portugal. His research focuses on transport demand modelling, intelligent transportation systems, data mining, machine learning, and the analysis of emerging mobility services and their integration with public transit systems.
\end{IEEEbiography}

\vspace{11pt}
\begin{IEEEbiography}[{\includegraphics[width=1in,height=1.25in,clip,keepaspectratio]{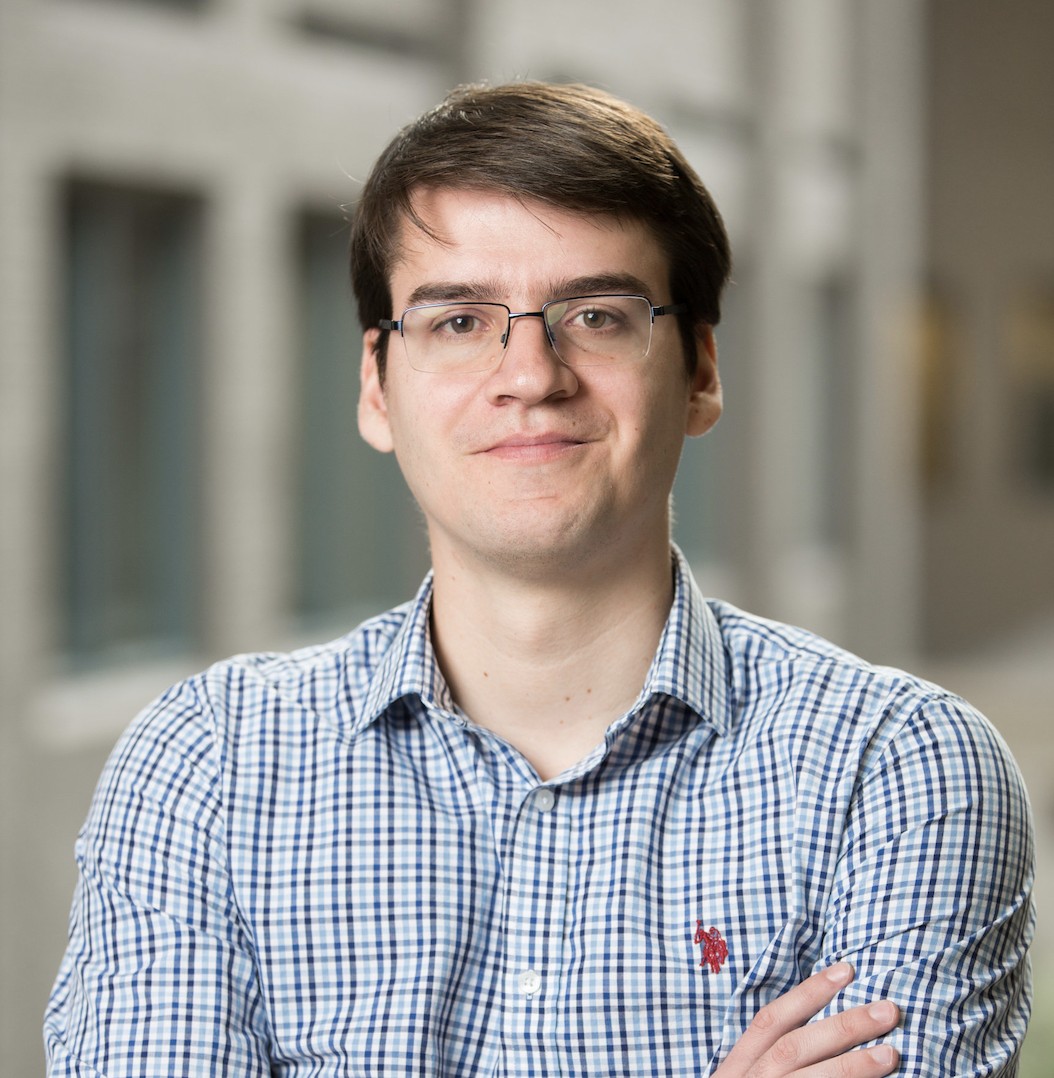}}]{Roberto Souza}
(Member, IEEE) received the B.Sc. degree in electrical engineering from the Federal University of Pará, Belém, Pará, Brazil, in 2012, and the M.Sc. and Ph.D. degrees in computer engineering from the University of Campinas (UNICAMP), Campinas, Brazil, in 2014 and 2017, respectively.

From 2017 to 2020, he was a Postdoctoral Scholar with the Department of Radiology, University of Calgary, Calgary, AB, Canada. He also held research internships with the Grenoble Institute of Technology, Grenoble, France, and the University of Pennsylvania, Philadelphia, PA, USA. In 2020, he joined the University of Calgary, where he is currently an Associate Professor with the Department of Electrical and Software Engineering, Calgary, AB, Canada. His research interests include image processing, machine learning, and data integration strategies for imaging applications, with a focus on computational methods for improving data mining, analysis, and interpretation in medical and scientific imaging.

Dr. Souza is a Registered Professional Engineer (P.Eng.) with the Association of Professional Engineers and Geoscientists of Alberta (APEGA).
\end{IEEEbiography}

\vfill

\end{document}